\documentclass[12pt,reqno]{amsart}
\usepackage{soul}
\usepackage{multicol,amsmath,graphics, mathtools, cancel,soul,color,bbm,amsfonts,dsfont,tikz,mathrsfs,amssymb,bm,bbold,comment}
\usepackage{algorithm}
\usepackage{algorithmicx}
\usepackage{algpseudocode}

\usepackage[foot]{amsaddr}

\usepackage{xcolor}
\usepackage[labelformat=parens,labelsep=space]{subcaption}

\usepackage{graphicx}

\usepackage{stmaryrd}
\usepackage[left=1in, right=1in, top=1.1in,bottom=1.1in]{geometry}
\usetikzlibrary{matrix,patterns,positioning}
\numberwithin{equation}{section}

\usepackage[normalem]{ulem} 
\newcommand{\mathsout}[1]{\ifmmode\text{\sout{\ensuremath{#1}}}\else\sout{#1}\fi}

\usepackage{comment}
\usepackage{soul}

\usepackage[hidelinks]{hyperref}

\newcommand{\vertiii}[1]{{\left\vert\kern-0.25ex\left\vert\kern-0.25ex\left\vert #1
    \right\vert\kern-0.25ex\right\vert\kern-0.25ex\right\vert}}

\def\dh2l{\mathbf{d}_{\mathbb{H}_{2\ell}}}
\def\d2{\mathbf{d}_2}

\newtheorem{theorem}{Theorem}[section]

\newtheorem{proposition}[theorem]{Proposition}

\theoremstyle{remark}

\theoremstyle{definition}

 \numberwithin{dummy}{section}

\def\1{\mathbbm{1}}

\makeatletter
\@namedef{subjclassname@1991}{2020 Mathematics Subject Classification}
\makeatother

\begin{document}

\title[Projected WFR Flows for Hawkes Background Estimation]{Projected Wasserstein-Fisher-Rao Flows for Hawkes Background-Intensity Estimation}
\author{Isa\'ias Ba\~nales}
\address{Facultad de Ciencias, Universidad Aut\'onoma de M\'exico}
\email{isaias@ciencias.unam.mx}

\author{Arturo Jaramillo}
\address{Department of Probability and Statistics, Centro de Investigaci\'on en Matem\'aticas A.C., Guanajuato, M\'exico}
\email{jagil@cimat.mx}

\begin{abstract}
We consider the problem of estimating the background intensity of a Hawkes process when the triggering mechanism is known, treating the background as an unknown finite positive measure with a regular density. We propose a likelihood-based variational approach based on the Wasserstein-Fisher-Rao geometry. In this setting, the relevant first variations are naturally distribution-valued and therefore fall outside the classical gradient-flow formulation. We develop a distributional framework for the resulting dynamics and give them a finite-dimensional meaning through projections onto statistical manifolds. We identify the true background as the unique zero of a limiting generalized gradient when $\kappa_{FR}>0$ and, under suitable stability, compactness, integrability and approximation conditions, establish a conditional consistency theorem for measurable estimators satisfying the empirical projected equations on events whose probabilities tend to one. Finally, we specialize the construction to Gaussian-mixture manifolds and obtain an explicit finite-dimensional scheme.\\

\noindent\textsc{Keywords:} Hawkes processes, ETAS models, background intensity estimation, Wasserstein-Fisher-Rao gradient flows, statistical manifolds, variational inference.
\end{abstract}

\maketitle

\section{Introduction}
This paper is devoted the to the development of a framework for variational inference based on distribution-valued Wasserstein-Fisher-Rao gradient dynamics on spaces of measures, with the goal of estimating the background intensity in Hawkes-type models of seismic activity. We also develop an explicit numerical scheme that exploits the merging of Gaussian components to obtain parsimonious representations of the background intensity. Its performance is illustrated through simulated Poisson data and real seismic data from southern Mexico, within the ETAS framework.\\

\noindent From the point of view of the statistical problem motivating this work, our starting point is the non-parametric estimation of the driving measures underlying point process models, which provide a natural framework for describing systems in which the occurrence of an event modifies the likelihood of subsequent events, and are particularly relevant in the modeling of seismic activity. Among the most prominent instances of such models are Hawkes processes \cite{Hawkes1971Spectra}, which can be viewed as counting processes retaining some of the basic features of the classical Poisson process, while incorporating a form of memory through a conditional intensity that combines a background component with a self-exciting contribution determined by the past history of the process (see Section \ref{subsec:hawkes-etas} for further details). In seismology, this mechanism constitutes the basis of the Epidemic Type Aftershock Sequence (ETAS) model \cite{ogata1988statistical}, where earthquakes are interpreted as events that may trigger additional seismic activity over time and space.\\

\noindent A fundamental statistical problem in this setting is the estimation of the background intensity. In a general non-parametric formulation, the background intensity is not represented by a finite-dimensional parameter, but by a non-negative measure, or equivalently, whenever absolute continuity holds, by a non-negative spatial density. The corresponding estimation problem is therefore naturally infinite dimensional. More precisely, given $N$ observed events with spatial locations $x_1,\ldots,x_N$ in a spatial region of interest $\mathcal Z\subseteq\mathbb R^d$ over a time horizon $[0,T]$, one is led to a likelihood functional $\mathcal F_T$ which, as explained in detail in Section~\ref{subsec:background-likelihood}, has the form
\begin{align}\label{eq:intro-likelihood}
\mathcal F_T(\mu)
=
\frac{1}{T}\sum_{q=1}^{N}
\log\big(
\Gamma[\mu](x_q)+b_q
\big)
-
\mu[\mathcal Z],
\end{align}
where $\mu$ denotes a finite nonnegative background measure on $\mathcal Z$ with a specified continuous density $\Gamma[\mu]$ with respect to Lebesgue measure, and $b_q$ represents the cumulative excitation at the $q$-th observed event generated by the preceding events. The triggering mechanism will be kept fixed throughout.\\

\noindent Maximizing $\mathcal F_T$ is a natural formulation of the estimation problem. Related likelihood-based optimization over spaces of measures using Wasserstein-Fisher-Rao dynamics has been studied by Yan, Wang and Rigollet \cite{yan2024gaussian}. Our setting, however, is different. We work directly with the Hawkes background likelihood as a functional of the unknown density, before introducing any finite-dimensional representation. As discussed in Section~\ref{subsec:background-likelihood}, the unrestricted likelihood maximization problem is ill posed. Nevertheless, the original likelihood still admits a distribution-valued first variation, structure that serves as the starting point for the projected dynamics developed in this work. To place the problem in context, we first review the broader literature on non-parametric background-intensity estimation for Hawkes and ETAS models, including likelihood-based, declustering, kernel, Bayesian and other non-parametric approaches. We then return to the variational formulation and explain how the ingredients used in our construction relate to existing work on measure-valued gradient flows and projection methods.

\subsection{Existing approaches to background intensity estimation}
\label{sec:Existing}
Among the existing approaches to background-intensity estimation, classical likelihood-based methods provide a natural starting point. In the classical ETAS formulation, the background and triggering mechanisms are specified through finite-dimensional parametric families and estimated by likelihood methods \cite{ogata1988statistical}. Zhuang, Ogata and Vere-Jones \cite{zhuang2002stochastic} introduced a stochastic declustering approach in which the branching structure is used to assign to each event a probability of belonging to the background process, whose intensity is then estimated non-parametrically through kernel methods. This latent branching interpretation was later used by Veen and Schoenberg \cite{veen2008estimation} to formulate an EM-type procedure, providing an alternative to the direct maximization of the likelihood; see also \cite{schoenberg2013facilitated}.\\

\noindent Several works have introduced more flexible representations of the background intensity. Marsan and Lenglin{\'e} \cite{marsan2008extending} considered a non-parametric reconstruction of earthquake triggering, while Fox, Schoenberg and Gordon \cite{fox2016spatially} developed spatially inhomogeneous background estimators for non-parametric Hawkes models. Li, Song and Wang \cite{li2020space} further considered space-time inhomogeneity by combining the latent branching representation with histogram and variable-bandwidth kernel estimators. These approaches allow considerable flexibility, but require choosing a particular representation or smoothing mechanism for the unknown intensity.\\

\noindent Bayesian non-parametric methods provide another important alternative. Kottas and Sans{\'o} \cite{kottas2007bayesian} represented the intensity of a spatial Poisson process as a positive scale multiplied by a probability density and assigned a Dirichlet-process mixture prior to the latter. Ross and Kolev \cite{ross2022semiparametric} incorporated this idea into the ETAS framework. Alternatively, Molkenthin et al.\ \cite{molkenthin2022gp} modeled the spatial background intensity through a Gaussian-process prior, while Muir and Ross \cite{muir2023deep} considered a deep Gaussian-process representation. Dependent Dirichlet-process constructions provide another way of allowing the underlying density to vary with space or time \cite{duan2007generalized,xiao2015modeling}. More recently, Ba\~nales et al.\ \cite{banales2026graphical} used graphical Dirichlet processes to model a spatiotemporally varying background intensity. Such flexibility is particularly relevant in seismic applications, where background activity may change due to tectonic forcing, slow-slip phenomena or large earthquakes \cite{hainzl2005detecting,hainzl2013impact,marsan2013monitoring,marsan2013slow,cruz2021short,nishikawa2023development}.

\subsection{Variational inference approach}\label{se:variationalinferenceapproach}

\noindent The variational formulation of likelihood-based estimation provides a natural connection with gradient flows on spaces of measures. This connection relies on a well-established mathematical framework \cite{jordan1998variational,otto2001geometry,ambrosio2008gradient}, and related constructions have been used in sampling and variational inference \cite{liu2016stein,liu2017stein,lambert2022variational}. In particular, Yan, Wang and Rigollet \cite{yan2024gaussian} develop Wasserstein-Fisher-Rao (WFR) gradient descent for non-parametric maximum-likelihood estimation of Gaussian mixtures. Their work provides an important methodological precedent for the construction considered here.\\

\noindent In the model studied by Yan, Wang and Rigollet \cite{yan2024gaussian}, independent and identically distributed observations have density $\xi_\star*\varphi$, where $\xi_\star$ is an unknown probability mixing measure and $\varphi$ is a fixed Gaussian density. The Gaussian convolution regularizes the dependence of the likelihood on the mixing measure, yielding a regular first variation with respect to this measure and allowing the Wasserstein Fisher Rao gradient flow to be formulated directly on the corresponding space of probability measures. The authors derive an interacting-particle implementation in which both weights and locations evolve, and establish convergence properties of the associated WFR iterations under suitable conditions.  Our construction, however, follows a different route: we derive the Wasserstein-Fisher-Rao direction directly from the original Hawkes likelihood at the level of the unknown background density, introducing a finite-dimensional representation only afterward through tangent-space projection.\\

\noindent In our setting, the unknown background intensity is a finite positive measure whose total mass must also be inferred. This naturally leads to unbalanced transport geometries \cite{kondratyev2016new,liero2018optimal,chizat2018unbalanced}, and in particular to the Wasserstein-Fisher-Rao geometry, whose Wasserstein component transports mass through space while its Fisher-Rao component allows its magnitude to vary.\\

\noindent When $\mu(dx)=\rho(x)\,dx$, the directional derivative of $\mathcal F_T$ with respect to regular enough perturbations of the density is represented by the distribution
\begin{align}\label{eq:intro-first-variation}
\frac{\delta\mathcal F_T}{\delta\rho}
=
\frac{1}{T}
\sum_{q=1}^{N}
\frac{1}{\rho(x_q)+b_q}
\delta_{x_q}
-
1.
\end{align}
This singular first variation prevents a direct use of the regular WFR gradient-flow formula, but it also provides the main computational simplification of our construction. When the corresponding WFR expression is tested against smooth directions, spatial derivatives are transferred onto those directions and the Dirac masses turn the empirical contribution into exact point evaluations at the observed events. We therefore retain the singular event terms rather than smoothing the empirical score. Once this weak action has been identified, we use the standard mechanism of tangent-space projection to obtain a finite-dimensional realization. The non-standard feature is thus not the projection step itself, but the distribution-valued Wasserstein Fisher Rao operator to which it is applied and the fact that its singular structure makes the resulting projected quantities explicit. Our main goal is to turn these dynamics into a practical likelihood-based procedure for estimating the background intensity, to study the limiting behavior of the resulting estimators, and to assess their performance numerically. We refer to the resulting construction as a projected Wasserstein-Fisher-Rao likelihood flow.\\

\noindent The use of tangent-space projection to obtain finite-dimensional approximations of infinite-dimensional evolutions is a well-established idea, closely related to Galerkin approximation and to the differential-geometric approach to statistical models. In nonlinear filtering, projection methods on finite-dimensional manifolds of densities were developed, in particular, by Brigo, Hanzon and Le Gland \cite{brigo1998projection,brigo1999approximate}; Armstrong and Brigo \cite{armstrong2016projectionL2} later developed direct-$L^2$ projection on Gaussian-mixture manifolds, with tangent Gram matrices, explicit Gaussian calculations and numerical implementations. Brigo and Pistone \cite{brigo2017measureprojection} place this type of projection-based dimensional reduction in a broader framework for measure-valued evolution equations. We use this established projection mechanism as one component of the present construction, applied here to the distribution-valued WFR operator generated by the Hawkes likelihood.\\

\subsubsection{Projected Wasserstein-Fisher-Rao likelihood flows}\mbox{}\par

\noindent Applying the Wasserstein-Fisher-Rao prescription to the empirical likelihood leads formally to $\partial_u\rho_u
=
\mathcal A_T(\rho_u)$, with 
\begin{align}\label{eq:intro-generalized-gradient}
\mathcal A_T(\rho)
=
-\kappa_W
\nabla\cdot\bigg(
\rho\nabla
\frac{\delta\mathcal F_T}{\delta\rho}
\bigg)
+
\kappa_{FR}
\rho
\frac{\delta\mathcal F_T}{\delta\rho}.
\end{align}
Since the first variation in \eqref{eq:intro-first-variation} is distribution-valued, we do not interpret this dynamic as a classical evolution equation. Instead, for a smooth finite-dimensional family
\[
\mathcal M_\Theta
=
\left\{
\rho_\theta:
\theta\in\Theta
\right\},
\]
with $\Theta\subset\mathbb R^m$ and $\rho_{\theta}$ a suitable family of densities, we require the tangent velocity to reproduce the action of $\mathcal A_T(\rho_\theta)$ on the tangent directions of $\mathcal M_\Theta$. The construction, developed in Section~\ref{sec:projected-formulation}, naturally leads to the determination of $\theta$ through the dynamics
\begin{align}\label{eq:intro-projected-flow}
G(\theta)\dot\theta
=
g_T(\theta),
\end{align}
where $G(\theta)$ is the Gram matrix of the tangent directions and $g_T(\theta)$ records the action of the generalized gradient on those directions. We call \eqref{eq:intro-projected-flow} the projected Wasserstein-Fisher-Rao likelihood flow.\\

\noindent This construction uses the likelihood to generate projected estimating equations; it does not in general produce likelihood ascent or likelihood maximization on the chosen manifold. Its statistical motivation is to ask whether empirical projected equilibria identify the true background as the observation horizon grows and the available tangent directions become sufficiently rich. This statistical limit is distinct from convergence of a trajectory in algorithmic time.

\subsubsection{Equilibrium}

\noindent Fix an observation horizon $T$. Under the regularity and immersion assumptions of Subsection~\ref{subsec:projected-equations}, suppose that a solution $\theta(u)$ of the empirical projected dynamics exists for every $u\geq0$ and converges to a parameter $\theta_\infty$ in $\Theta$. Then $\rho_\Theta^\circ:=\rho_{\theta_\infty}$ satisfies the empirical equilibrium equation
\begin{align}\label{eq:Amuodrel}
\left\langle
\mathcal A_T(\rho_\Theta^\circ),
d
\right\rangle
=
0
\end{align}
for every $d$ in $ T_{\rho_\Theta^\circ}\mathcal M_\Theta$. Thus, each statistical manifold provides a finite collection of empirical projected equations. This implication does not establish convergence of the trajectories.\\

\subsubsection{Conditional consistency}

\noindent In order to carry out the consistency analysis, we assume that the data-generating Hawkes-ETAS parameters satisfy the conditions stated in Section~\ref{subsec:hawkes-etas}, including the subcriticality condition on the branching ratio. The precise arguments are developed in Section~\ref{sec:consistency}; here we only describe the mechanism behind the result.\\

\noindent Let $\rho_\star$ denote the true background density and let $\rho$ be a generic candidate density, with the triggering mechanism kept fixed. The observed data are generated under $\rho_\star$, while the empirical likelihood is evaluated at $\rho$. Applying the projected Wasserstein-Fisher-Rao construction to this likelihood gives the empirical generalized gradient $\mathcal A_T(\rho)$. The consistency question is whether the equilibria produced by this procedure identify the background density that generated the data. To address this, we introduce the limiting generalized gradient $\mathcal A(\rho)$ associated with the stationary regime of the model. As a first step, for $\kappa_{FR}>0$, we prove that $\mathcal A(\rho)=0$ if and only if $\rho=\rho_\star$. Thus, at the limiting level, the generalized-gradient equation has the true background density as its unique solution.\\

\noindent It remains to understand whether the empirical projected equilibria are eventually forced toward this limiting equation. Two approximations are involved, and they play different roles. First, as the observation horizon grows, the empirical generalized gradient $\mathcal A_T(\rho)$ approaches its limiting counterpart $\mathcal A(\rho)$ in a suitable topology of distributions. This is where the stationary and ergodic regime provided by the subcriticality assumptions enters the argument. Second, as the statistical manifolds become sufficiently rich, their tangent spaces provide an increasingly complete collection of directions along which the generalized gradient is tested. Hence, if $\rho_n^\circ$ is an empirical projected equilibrium satisfying
\[
\left\langle
\mathcal A_{T_n}(\rho_n^\circ),
v
\right\rangle
=
0,
\qquad
v\in T_{\rho_n^\circ}\mathcal M_{m_n},
\]
then, for large observation horizons and sufficiently accurate tangent approximations, these projected equations increasingly capture the full limiting condition $\mathcal A(\rho)=0$. Since that condition identifies only $\rho_\star$, this suggests that the empirical equilibria should approach the true background density.\\

\noindent Theorem~\ref{thm:projected-consistency} makes this argument precise under the assumptions of Section~\ref{sec:consistency}, including tangent approximation and the high-probability existence of measurable empirical equilibria. The selected densities then converge in probability to $\rho_\star$ in both $H^\beta(\mathcal Z)$ and $L^1(\mathcal Z)$, and their total masses converge to that of $\rho_\star$.

\subsection{Gaussian-mixture realization and numerical study}

\noindent We finally turn to a concrete realization of the projected construction. For the numerical implementation, we take the statistical manifolds to be finite mixtures of Gaussian densities with fixed bandwidth. For a given number $k$ of components, we write
\begin{align}\label{eq:intro-gaussian-mixture}
\rho_\theta(x)
=
\sum_{i=1}^k
e^{\alpha_i}\chi_\sigma(x;m_i),
\end{align}
where $m_1,\ldots,m_k$ are the component locations and $\alpha_1,\ldots,\alpha_k$ are their log-weights. This choice is particularly convenient for the equations developed above: the densities and their tangent directions are smooth, the Gram matrix can be computed explicitly, and the action of the distributional generalized gradient on the tangent space admits a closed expression in terms of the observed events, the Gaussian components and the Hawkes excitation. The resulting projected flow is therefore a finite-dimensional system of ordinary differential equations that can be implemented without discretizing the ambient spatial domain. \\

\noindent The number of components $k$ controls the flexibility of the finite-dimensional Gaussian representation and the number of tangent directions available to the projected equations. In the planned numerical experiments, $k$ will be treated as an implementation parameter rather than as the index of an asymptotic manifold-refinement scheme. Thus, varying $k$ allows us to examine the effect of the dimension of the chosen statistical manifold, without invoking the tangent-approximation assumptions of Theorem~\ref{thm:projected-consistency}.\\

\noindent The planned numerical experiments will examine the resulting dynamics in examples for which the true background intensity is known. We will study how well the method recovers spatially non-constant background intensities and how the reconstruction changes with the number of Gaussian components, their initialization and bandwidth, and the relative values of $\kappa_W$ and $\kappa_{FR}$. In particular, the experiments will examine the contributions inherited from the ambient Wasserstein transport and Fisher-Rao reaction terms.\\

\noindent Section~\ref{sec:preliminaries} introduces the model and background likelihood. Section~\ref{sec:projected-formulation} develops the distributional WFR construction and its projection onto statistical manifolds. Section~\ref{sec:gaussian-mixtures} gives the Gaussian-mixture realization. Section~\ref{sec:consistency} presents a conditional consistency result for an increasingly complex collection of statistical manifolds. Numerical experiments and discussion follow in Sections~\ref{sec:numerical-experiments} and \ref{sec:discussion}. The appendices contain the statistical proofs.


\section{Model and background likelihood}\label{sec:preliminaries}

\noindent We first specify the observed process and isolate the part of its likelihood that depends on the background. The triggering mechanism is kept fixed throughout, so changes in the candidate background do not change the excitation computed from a given event history.

\subsection{Hawkes processes and the ETAS model}\label{subsec:hawkes-etas}

\noindent Let $\boldsymbol\eta$ be a locally finite simple marked point process on $\mathbb R_+\times\mathcal Z\times\mathcal R$, where $\mathcal Z\subseteq\mathbb R^d$ is the spatial region and $\mathcal R$ is the mark space. Its ground counting measure is $\eta(dt,dx):=\boldsymbol\eta(dt,dx,\mathcal R)$. We use the usual augmentation of the natural filtration and start with empty history at time zero. Under a background density $\rho$, the ground process has predictable compensator $\lambda_\rho(t,x)\,dt\,dx$. Equivalently, for each bounded Borel set $C\subseteq\mathcal Z$,
\[
\eta[(0,t]\times C]-\int_0^t\int_C\lambda_\rho(s,x)\,dx\,ds
\]
is a local martingale. We take
\begin{align}\label{eq:hawkes-background-decomposition}
\lambda_\rho(t,x)&=\rho(x)+B_t(x),\\
B_t(x)&=\int_{(0,t)}\int_{\mathcal Z\times\mathcal R}\nu(t-s,x-y,r)\,\boldsymbol\eta(ds,dy,dr).\label{eq:hawkes-excitation-field}
\end{align}
The background is time-independent, nonnegative and has finite mass. The kernel $\nu$ is nonnegative, and $B_t$ only uses events strictly before $t$. For finitely many previous events, $B_t(x)=\sum_{i:t_i<t}\nu(t-t_i,x-x_i,r_i)$. Thus $\rho$ describes background activity, while the summands describe activity triggered by earlier events. General references are \cite{Hawkes1971Spectra,daley2003introduction,hawkes1974cluster,bremaud1996stability}.

\noindent For the ETAS specification \cite{ogata1988statistical}, use the factorization
\begin{align}\label{eq:etas-kernel-schematic}
\nu(s,z,r)=\kappa(r)g(s)f(z;r)\mathbbm 1_{\{s>0\}},
\end{align}
where $\kappa$ gives productivity, $g$ gives the temporal distribution of direct descendants, and $f(\cdot;r)$ is a probability density on $\mathbb R^d$. More specifically, let
\[
\kappa(r)=K\exp\{\alpha(r-M_u)\},\qquad g(s)=(p-1)c^{p-1}(s+c)^{-p},\qquad s>0,
\]
with $K,c>0$ and $p>1$, so that $\int_0^\infty g(s)\,ds=1$. Marks attached at birth are independent of the preceding history and of one another, with density
\begin{align}\label{eq:gr-mark-density}
q_{\mathrm{GR}}(r)=\beta_{\mathrm{GR}}\exp\{-\beta_{\mathrm{GR}}(r-M_u)\}\mathbbm 1_{\{r\geq M_u\}},\qquad\beta_{\mathrm{GR}}>0.
\end{align}
The marked compensator is therefore $\lambda_\rho(t,x)q_{\mathrm{GR}}(r)\,dt\,dx\,dr$.

\noindent The finite-sample construction does not require stationarity. For the statistical analysis, we assume $\alpha<\beta_{\mathrm{GR}}$ and
\begin{align}\label{eq:etas-subcriticality}
\mathfrak n:=\int_{M_u}^\infty\kappa(r)q_{\mathrm{GR}}(r)\,dr=K\frac{\beta_{\mathrm{GR}}}{\beta_{\mathrm{GR}}-\alpha}<1.
\end{align}
On $\mathbb R^d$, $\mathfrak n$ is the expected number of direct descendants after averaging over the parent mark. On a smaller region, discarding descendants outside the region can only decrease this mean. Finite background mass and \eqref{eq:etas-subcriticality} give a stationary version with finite mean rate that is mixing under time translations. Appendix~\ref{app:uniform-law-large-numbers} recalls the construction and gives the estimates needed to compare it with the observed empty-history process. No spatial stationarity is assumed.

\subsection{Likelihood for the background intensity}\label{subsec:background-likelihood}

\noindent Let $(t_q,x_q,r_q)_{q=1}^N$ be the events observed up to $T>0$, ordered by time, and put $b_q:=B_{t_q}(x_q)$. The joint marked point-process likelihood is
\begin{align}\label{eq:conditional-intensity-likelihood}
L_T^{\mathrm{marked}}(\rho)=\exp\bigg\{-\int_0^T\int_{\mathcal Z}\lambda_\rho(t,x)\,dx\,dt\bigg\}\prod_{q=1}^N\lambda_\rho(t_q,x_q)q_{\mathrm{GR}}(r_q).
\end{align}
The intensity is evaluated predictably, so an event does not contribute to its own excitation. The mark-density factor is independent of the candidate background and can be omitted when estimating $\rho$; this is a factorization of the joint likelihood, not conditioning on the entire observed list of marks. See \cite{daley2003introduction,ogata1988statistical}.

\noindent Since the triggering mechanism is fixed, $B_t$ is computed from the same observed history for every candidate. In particular,
\[
\int_0^T\int_{\mathcal Z}\lambda_\rho(t,x)\,dx\,dt=T\int_{\mathcal Z}\rho(x)\,dx+\int_0^T\int_{\mathcal Z}B_t(x)\,dx\,dt.
\]
The log-likelihood can consequently be written as
\begin{align}\label{eq:background-log-likelihood-reduced}
\ell_T(\rho)=\sum_{q=1}^N\log\big(\rho(x_q)+b_q\big)-T\int_{\mathcal Z}\rho(x)\,dx+C_T,
\end{align}
where $C_T$ includes the integrated excitation and mark contribution and does not depend on $\rho$. We therefore work with the normalized objective
\begin{align}\label{eq:normalized-empirical-likelihood}
\mathcal F_T(\rho)=\frac1T\sum_{q=1}^N\log\big(\rho(x_q)+b_q\big)-\int_{\mathcal Z}\rho(x)\,dx.
\end{align}
For $\mu(dx)=\rho(x)\,dx$, this is the functional in \eqref{eq:intro-likelihood}. Normalization rescales the subsequent vector field and algorithmic time, but not its equilibria.

\noindent The point evaluations require specified continuous representatives. For finite observations, we initially use integrable densities with $\rho(x_q)+b_q>0$ at every event and finite $b_q$; the construction below imposes additional spatial regularity. This restriction matters statistically: unrestricted concentration at observed points can make the likelihood arbitrarily large while keeping the mass bounded. We do not assume an unrestricted non-parametric likelihood maximizer exists. Instead, the likelihood generates projected estimating equations on smooth statistical manifolds.

\section{Projected Wasserstein-Fisher-Rao construction}\label{sec:projected-formulation}\label{sec:variational-formulation}

\noindent In this section we make precise the dynamics of measures outlined in the introduction. The construction proceeds in four steps. First, we recall how the Wasserstein-Fisher-Rao geometry associates an evolution equation with a functional whose first variation is regular. Second, we return to the empirical likelihood: since its first variation in \eqref{eq:intro-first-variation} is distribution-valued, we interpret the corresponding WFR gradient in a weak sense and identify a Sobolev setting in which the resulting generalized gradient $\mathcal A_T(\rho)$ is well defined. Third, we establish the regularity of the map $\rho\mapsto\mathcal A_T(\rho)$ that will be needed to define the dynamics. Finally, having identified the generalized WFR gradient that prescribes the infinitesimal motion of the measure, we use it to define dynamics on smooth finite-dimensional statistical manifolds. Since this ambient direction is distribution-valued and need not itself be a tangent vector to the chosen manifold, we select the tangent velocity whose $L^2$ pairings with the tangent directions reproduce the action prescribed by the generalized WFR gradient. Each statistical manifold therefore yields a finite-dimensional projected realization of the same underlying infinitesimal WFR dynamics, and the collection of these realizations constitutes our projected Wasserstein-Fisher-Rao construction.\\

\noindent Throughout this section and the remaining analysis, we take $\mathcal Z=\mathbb R^d$. The observation horizon is $T$, while $u$ denotes algorithmic time.\\

\subsection{Wasserstein Fisher Rao dynamics}\label{subsec:regular-wfr}

\noindent We begin by recalling the standard Wasserstein-Fisher-Rao formulation underlying our construction; see \cite{jordan1998variational,otto2001geometry,ambrosio2008gradient,santambrogio2015optimal} for the Wasserstein gradient-flow viewpoint, and \cite{kondratyev2016new,liero2018optimal,chizat2018unbalanced,chewi2025statistical} for the Wasserstein-Fisher-Rao geometry and its applications to gradient dynamics. For a smooth family $\rho_u$ of positive, integrable densities on $\mathcal Z$, a transport field $v_u$ and a reaction field $\alpha_u$ represent its velocity through
\begin{align}\label{eq:wfr-continuity-reaction}
\partial_u\rho_u+\nabla\cdot(\rho_u v_u)=\alpha_u\rho_u.
\end{align}
The pair $(v_u,\alpha_u)$ is not uniquely determined by the velocity $\partial_u\rho_u$: different combinations of transport and reaction may produce the same change of density through \eqref{eq:wfr-continuity-reaction}. To distinguish among these representations, one equips the space of measures with a metric structure, or equivalently at the infinitesimal level with a cost for the pair $(v_u,\alpha_u)$. Minimizing this cost among all pairs producing the same velocity defines the corresponding tangent norm. Thus, \eqref{eq:wfr-continuity-reaction} specifies the admissible ways in which a measure may evolve, while the choice of cost determines the geometry used to select among them. In the WFR geometry, for $\kappa_W,\kappa_{FR}>0$, this cost takes the form
\[
\int_{\mathcal Z}\bigg(\frac{|v_u|^2}{\kappa_W}+\frac{|\alpha_u|^2}{\kappa_{FR}}\bigg)\rho_u\,dx.
\]
The squared WFR norm of the tangent velocity is then defined as the smallest cost among all pairs representing it:
\begin{align}\label{eq:wfr-tangent-norm}
\|\partial_u\rho_u\|_{\rho_u,\mathrm{WFR}}^2
=
\inf_{\substack{v_u,\alpha_u:\\ \partial_u\rho_u+\nabla\cdot(\rho_u v_u)=\alpha_u\rho_u}}
\int_{\mathcal Z}\bigg(\frac{|v_u|^2}{\kappa_W}+\frac{|\alpha_u|^2}{\kappa_{FR}}\bigg)\rho_u\,dx.
\end{align}
The operator formulas below also allow nonnegative coefficients with $\kappa_W+\kappa_{FR}>0$; the displayed tangent norm is used only when both coefficients are positive.

\noindent To see how this geometry determines a gradient, suppose a functional $\mathcal E$ has a regular first variation $\varphi_\rho$. This means that
\[
D\mathcal E(\rho)[h]:=\left.\frac{d}{d\varepsilon}\mathcal E(\rho+\varepsilon h)\right|_{\varepsilon=0}=\int_{\mathcal Z}\varphi_\rho(x)h(x)\,dx
\]
for admissible perturbations $h$. Along a curve satisfying \eqref{eq:wfr-continuity-reaction}, the chain rule and integration by parts give
\begin{align}\label{eq:wfr-differential}
\frac{d}{du}\mathcal E(\rho_u)&=\int_{\mathcal Z}\varphi_{\rho_u}\,\partial_u\rho_u\,dx
=\int_{\mathcal Z}\rho_u\nabla\varphi_{\rho_u}\cdot v_u\,dx+\int_{\mathcal Z}\rho_u\varphi_{\rho_u}\alpha_u\,dx.
\end{align}
With respect to the weighted inner product in \eqref{eq:wfr-tangent-norm}, this derivative is represented by the fields $v_u=\kappa_W\nabla\varphi_{\rho_u}$ and $\alpha_u=\kappa_{FR}\varphi_{\rho_u}$. Substitution into the continuity-reaction equation gives the formal WFR gradient-ascent equation
\begin{align}\label{eq:wfr-formal-ascent}
\partial_u\rho_u=-\kappa_W\nabla\cdot(\rho_u\nabla\varphi_{\rho_u})+\kappa_{FR}\rho_u\varphi_{\rho_u}.
\end{align}
Along this regular evolution,
\[
\frac{d}{du}\mathcal E(\rho_u)=\kappa_W\int_{\mathcal Z}\rho_u|\nabla\varphi_{\rho_u}|^2\,dx+\kappa_{FR}\int_{\mathcal Z}\rho_u|\varphi_{\rho_u}|^2\,dx\geq0.
\]
Thus, whenever the preceding formal calculation is justified, the WFR gradient direction increases the value of $\mathcal E$ along the resulting curve.\\

\subsection{Distributional Wasserstein Fisher Rao gradient}\label{subsec:empirical-generalized-gradient}
\noindent We now return to the empirical likelihood. Fix the observed events and write $b_q=B_{t_q}(x_q)\in[0,\infty)$. Let $\rho\in C^1(\mathcal Z)\cap L^1(\mathcal Z)$ be nonnegative and satisfy $\rho(x_q)+b_q>0$ at every event. Differentiating \eqref{eq:normalized-empirical-likelihood} in an integrable, continuous direction $h$ for which $\rho+\varepsilon h$ remains admissible gives
\begin{align}\label{eq:empirical-directional-derivative}
D\mathcal F_T(\rho)[h]=\frac1T\sum_{q=1}^N\frac{h(x_q)}{\rho(x_q)+b_q}-\int_{\mathcal Z}h(x)\,dx.
\end{align}
Using $\langle\delta_{x_q},h\rangle=h(x_q)$, the directional derivative can be written as
\[
D\mathcal F_T(\rho)[h]=\langle\mathcal S_T(\rho),h\rangle,
\]
thus implying that the associated first variation is represented by
\begin{align}\label{eq:empirical-score-distribution}
\mathcal S_T(\rho)=\frac1T\sum_{q=1}^N\frac{\delta_{x_q}}{\rho(x_q)+b_q}-1.
\end{align}
Here the constant term acts by integration against Lebesgue measure.

\noindent If we formally substitute $\mathcal S_T(\rho)$ into the WFR equation \eqref{eq:wfr-formal-ascent}, the Wasserstein term would require taking spatial derivatives of the Dirac masses appearing in $\mathcal S_T(\rho)$. This suggests formulating the corresponding WFR gradient in a weak sense. First suppose that the first variation is represented by a smooth function $s$. For $\phi\in C_c^\infty(\mathcal Z)$, integration by parts gives
\begin{align*}
&\int_{\mathcal Z}\big[-\kappa_W\nabla\cdot(\rho\nabla s)+\kappa_{FR}\rho s\big]\phi\,dx\\
&\qquad=-\kappa_W\int_{\mathcal Z}s\nabla\cdot(\rho\nabla\phi)\,dx+\kappa_{FR}\int_{\mathcal Z}s\rho\phi\,dx.
\end{align*}
The right-hand side no longer involves derivatives of $s$ and therefore remains meaningful when $s$ is replaced by the distributional first variation $\mathcal S_T(\rho)$. This leads to the following weak formulation of the empirical WFR generalized gradient:
\begin{align}\label{eq:empirical-wfr-force-weak}
\langle\mathcal A_T(\rho),\phi\rangle
:=
-\kappa_W\langle\mathcal S_T(\rho),\nabla\cdot(\rho\nabla\phi)\rangle
+\kappa_{FR}\langle\mathcal S_T(\rho),\rho\phi\rangle.
\end{align}
The pairings are understood as integration against the locally finite signed measure $\mathcal S_T(\rho)$; their arguments are continuous and compactly supported.

\noindent The weak formulation above can be written explicitly in terms of the observed events. Substituting \eqref{eq:empirical-score-distribution} into \eqref{eq:empirical-wfr-force-weak} and using
\[
\int_{\mathcal Z}\nabla\cdot(\rho\nabla\phi)\,dx=0,
\]
for $\phi\in C_c^\infty(\mathcal Z)$, gives
\begin{align}\label{eq:empirical-wfr-force-explicit}
\langle\mathcal A_T(\rho),\phi\rangle
={}&-\frac{\kappa_W}{T}\sum_{q=1}^N\frac{\nabla\cdot(\rho\nabla\phi)(x_q)}{\rho(x_q)+b_q}
+\frac{\kappa_{FR}}{T}\sum_{q=1}^N\frac{\rho(x_q)\phi(x_q)}{\rho(x_q)+b_q}\nonumber\\
&-\kappa_{FR}\int_{\mathcal Z}\rho(x)\phi(x)\,dx.
\end{align}

\subsection{Sobolev regularity}\label{subsec:empirical-gradient-regularity}
\noindent Before passing to statistical manifolds, we need to make sure that the weak action above is defined on a sufficiently large class of test functions and depends regularly on the candidate density. This is precisely what will allow us to evaluate $\mathcal A_T(\rho)$ on tangent directions and, later, to obtain a well-defined finite-dimensional dynamics. Since \eqref{eq:empirical-wfr-force-explicit} involves pointwise evaluations of $\phi$, $\nabla\phi$ and $\Delta\phi$, we take
\[
\beta>\frac d2+2.
\]
Then Sobolev embedding gives a constant $C_\beta$ such that
\[
\max\{\|\phi\|_\infty,\|\nabla\phi\|_\infty,\|\Delta\phi\|_\infty\}
\leq C_\beta\|\phi\|_{H^\beta}.
\]
We write $H^{-\beta}$ for the continuous dual of $H^\beta$. The following result shows that the weak empirical WFR gradient defines an element of this dual space, and hence in particular a tempered distribution, and varies regularly with $\rho$.

\begin{proposition}\label{prop:empirical-generalized-gradient}
Let $\beta>d/2+2$ and let $\rho\in C^1(\mathcal Z)\cap L^1(\mathcal Z)$ be nonnegative, with $\rho(x_q)+b_q>0$ for every observed event. Then the action defined in \eqref{eq:empirical-wfr-force-weak} extends uniquely from $C_c^\infty(\mathcal Z)$ to a continuous linear functional on $H^\beta$. In particular,
\[
\mathcal A_T(\rho)\in H^{-\beta}.
\]
For every $\phi\in H^\beta$, its action is given by \eqref{eq:empirical-wfr-force-explicit}. Moreover, for every $M,a>0$, there is a finite constant $C$, depending also on the fixed observations, $T$, $\beta$ and the coefficients, such that
\[
\|\mathcal A_T(\rho)-\mathcal A_T(\gamma)\|_{H^{-\beta}}\leq C\|\rho-\gamma\|_{H^\beta}
\]
whenever $\rho,\gamma\in H^\beta\cap L^1$ are nonnegative, their $H^\beta$ norms are at most $M$, and $\rho(x_q)+b_q,\gamma(x_q)+b_q\geq a$ at every event.
\end{proposition}

\begin{proof}
For $\phi\in C_c^\infty(\mathcal Z)$, formula \eqref{eq:empirical-wfr-force-explicit} and Sobolev embedding give
\begin{align*}
|\langle\mathcal A_T(\rho),\phi\rangle|
\leq{}&\frac{C_\beta}{T}\sum_{q=1}^N\frac{\kappa_W|\nabla\rho(x_q)|+(\kappa_W+\kappa_{FR})\rho(x_q)}{\rho(x_q)+b_q}\|\phi\|_{H^\beta}\\
&+\kappa_{FR}C_\beta\|\rho\|_{L^1}\|\phi\|_{H^\beta}.
\end{align*}
The coefficient is finite because the number of observed events is finite and every denominator is positive. Hence the weak action is bounded with respect to the $H^\beta$ norm on $C_c^\infty(\mathcal Z)$. Density of $C_c^\infty(\mathcal Z)$ in $H^\beta$ gives a unique continuous extension, so $\mathcal A_T(\rho)\in H^{-\beta}$. The right-hand side of \eqref{eq:empirical-wfr-force-explicit} is itself continuous in $\phi$ for the $H^\beta$ norm by the same bound, and therefore represents this extension.\\

\noindent For the Lipschitz property, let $\rho$ and $\gamma$ have $H^\beta$ norms bounded by $M$ and suppose that
\[
\rho(x_q)+b_q,\ \gamma(x_q)+b_q\geq a>0
\]
for every observed event. At each $x_q$, the difference of the Wasserstein contributions can be written as
\begin{align*}
\frac{\nabla\cdot(\rho\nabla\phi)}{\rho+b_q}
-\frac{\nabla\cdot(\gamma\nabla\phi)}{\gamma+b_q}
={}&
\frac{\nabla\cdot((\rho-\gamma)\nabla\phi)}{\rho+b_q}
+\nabla\cdot(\gamma\nabla\phi)\frac{\gamma-\rho}{(\rho+b_q)(\gamma+b_q)}.
\end{align*}
The same decomposition applies to the Fisher-Rao event term. Sobolev embedding and the lower bound $a$ bound each event difference by a constant times $\|\rho-\gamma\|_{H^\beta}\|\phi\|_{H^\beta}$. The deterministic term is controlled separately by
\[
\bigg|\int_{\mathcal Z}(\rho-\gamma)\phi\,dx\bigg|\leq\|\rho-\gamma\|_{L^2}\|\phi\|_{L^2}\leq\|\rho-\gamma\|_{H^\beta}\|\phi\|_{H^\beta}.
\]
Combining these bounds gives
\[
|\langle\mathcal A_T(\rho)-\mathcal A_T(\gamma),\phi\rangle|
\leq
C\|\rho-\gamma\|_{H^\beta}\|\phi\|_{H^\beta},
\]
where $C$ depends on $M,a,T,\beta$, the coefficients and the fixed observations. Taking the supremum over $\|\phi\|_{H^\beta}\leq1$ proves the claim.
\end{proof}

\noindent For the statistical analysis developed later, it will also be convenient to write the same action directly in terms of the ground counting measure:
\begin{align}\label{eq:empirical-generalized-gradient-integral}
\langle\mathcal A_T(\rho),\phi\rangle
={}&\frac1T\int_{(0,T]\times\mathcal Z}
\frac{-\kappa_W\nabla\cdot(\rho\nabla\phi)(x)+\kappa_{FR}\rho(x)\phi(x)}
{\rho(x)+B_t(x)}
\,\eta(dt,dx)\nonumber\\
&-\kappa_{FR}\int_{\mathcal Z}\rho(x)\phi(x)\,dx.
\end{align}
Thus $\mathcal A_T(\rho)$ can be evaluated on any $H^\beta$ tangent direction, which is exactly what is needed to define the projected dynamics in the next subsection.

\subsection{Projected dynamics}\label{subsec:projected-equations}

\noindent Having constructed the generalized WFR gradient $\mathcal A_T(\rho)$, which prescribes the ambient infinitesimal motion of the density, we now use it to define a finite-dimensional dynamics. For a chosen statistical manifold, the ambient distributional direction need not itself be a tangent vector. We therefore determine a tangent velocity by requiring its $L^2$ pairing with each tangent direction to reproduce the action of $\mathcal A_T(\rho)$ on that direction. Let $\Theta\subset\mathbb R^m$ be open and let $\theta\mapsto\rho_\theta$ be twice continuously differentiable into $H^\beta$. Each $\rho_\theta$ is an integrable nonnegative density satisfying $\rho_\theta(x_q)+b_q>0$ at every observed event. Write
\begin{align}\label{eq:tangent-directions}
\mathcal M_\Theta=\{\rho_\theta:\theta\in\Theta\},\qquad d_i^\theta=\partial_{\theta_i}\rho_\theta,\qquad T_{\rho_\theta}\mathcal M_\Theta=\operatorname{span}\{d_1^\theta,\ldots,d_m^\theta\}.
\end{align}
We assume the parametrization is an immersion, meaning that these tangent directions are linearly independent in $L^2$. We work in local charts of an embedded statistical manifold; redundant parameter labels are permitted only when they are related by compatible changes of coordinates. A differentiable parameter curve satisfies $\partial_u\rho_{\theta(u)}=\sum_i d_i^{\theta(u)}\dot\theta_i(u)$. The empirical projected equations are
\begin{align}\label{eq:projected-weak-condition}
\int_{\mathcal Z}\partial_u\rho_{\theta(u)}(x)d_j^{\theta(u)}(x)\,dx=\langle\mathcal A_T(\rho_{\theta(u)}),d_j^{\theta(u)}\rangle,
\end{align}
for $j=1,\ldots,m$. They replace an equality between a regular tangent velocity and a distribution by finitely many scalar equalities.\\

\noindent Define the Gram matrix and the projected coefficients by
\begin{align}\label{eq:projected-gram}
G_{ij}(\theta)=\int_{\mathcal Z}d_i^\theta d_j^\theta\,dx,\qquad g_{T,j}(\theta)=\langle\mathcal A_T(\rho_\theta),d_j^\theta\rangle.
\end{align}
The chain rule turns \eqref{eq:projected-weak-condition} into
\begin{align}\label{eq:projected-ode}
G(\theta)\dot\theta=g_T(\theta),\qquad \dot\theta=G(\theta)^{-1}g_T(\theta).
\end{align}
Indeed, $a^\top G(\theta)a=\|\sum_i a_i d_i^\theta\|_{L^2}^2$, so the immersion assumption makes $G$ positive definite. Substituting \eqref{eq:empirical-wfr-force-explicit} gives
\begin{align}\label{eq:projected-force-explicit}
g_{T,j}(\theta)
={}&-\frac{\kappa_W}{T}\sum_{q=1}^N\frac{\nabla\cdot(\rho_\theta\nabla d_j^\theta)(x_q)}{\rho_\theta(x_q)+b_q}
+\frac{\kappa_{FR}}{T}\sum_{q=1}^N\frac{\rho_\theta(x_q)d_j^\theta(x_q)}{\rho_\theta(x_q)+b_q}\nonumber\\
&-\kappa_{FR}\int_{\mathcal Z}\rho_\theta d_j^\theta\,dx.
\end{align}

\noindent Equations \eqref{eq:projected-weak-condition} identify the unique tangent vector whose $L^2$ pairing with every tangent direction agrees with the action of $\mathcal A_T$. Thus the velocity is the finite-dimensional $L^2$-Riesz representative of the restricted functional, not an $L^2$ representative of the ambient distribution. This description is independent of coordinates: a smooth reparametrization changes the matrix and vector but not the tangent velocity of the density. It is not the gradient obtained by restricting the WFR metric to $\mathcal M_\Theta$, so neither likelihood ascent nor likelihood maximization on the manifold is implied.\\

\noindent On a sufficiently small coordinate neighborhood, the event intensities have a positive common lower bound and the parametrization and its first two derivatives have bounded $H^\beta$ norms. Proposition~\ref{prop:empirical-generalized-gradient} and the continuous bilinear dual pairing then make $g_T$ locally Lipschitz. Since $G$ is continuously differentiable and positive definite, $G^{-1}g_T$ is locally Lipschitz. Hence, for every initial condition $\theta_0\in\Theta$, the projected system \eqref{eq:projected-ode} admits a unique local solution by standard ordinary differential equation theory.\\

\noindent The same construction applies to every statistical manifold satisfying the preceding assumptions. The generalized WFR gradient $\mathcal A_T$ prescribes the underlying ambient infinitesimal motion, while the chosen manifold determines its finite-dimensional tangent realization. The resulting collection of manifold-indexed systems \eqref{eq:projected-ode} is what we call the projected Wasserstein-Fisher-Rao dynamics.\\

\noindent An empirical projected equilibrium satisfies $g_T(\theta)=0$. If a global trajectory has an interior limit $\theta_\infty$, continuity of the vector field makes its velocity converge to $G(\theta_\infty)^{-1}g_T(\theta_\infty)$. That vector must be zero: otherwise its projection along its limiting direction is eventually bounded away from zero, contradicting convergence of $\theta(u)$. Existence of a trajectory limit is not asserted. The statistical question in Section~\ref{sec:consistency} instead concerns empirical equilibria as the observation horizon and the collection of available tangent directions increase.

\section{Gaussian-mixture realization}\label{sec:gaussian-mixtures}

\noindent In the previous section, the projected Wasserstein-Fisher-Rao construction was defined for a general statistical manifold, giving a family of finite-dimensional realizations of the same ambient generalized WFR dynamics. Our purpose here is not to study all such realizations simultaneously, but to obtain one explicit and computable instance that can be used for estimation. We therefore specialize to fixed-bandwidth Gaussian mixtures. For this family, all the ingredients of the projected system introduced in \eqref{eq:intro-projected-flow} can be computed explicitly: we first identify the tangent directions, then compute their $L^2$ Gram matrix, next evaluate the generalized WFR gradient along these directions, and finally assemble the resulting system in a form suitable for numerical implementation. The statistical conditions under which projected estimators recover the true background are studied in the forthcoming Section~\ref{sec:consistency}.\\

\subsection{Tangent directions}\label{subsec:gaussian-tangent-directions}

\noindent The first ingredient to be determined is the tangent space of the chosen statistical manifold.  Fix $k\geq1$ and $\sigma>0$. We write
\[
\theta=(\alpha_1,\ldots,\alpha_k,m_1,\ldots,m_k)
\]
for the parameters of the Gaussian mixture, where $\alpha_i\in\mathbb R$ are the log-weights and $m_i\in\mathbb R^d$ are the component locations. We then define
\begin{align}\label{eq:gaussian-kernel}
\chi_\sigma(x;m)&=(2\pi\sigma^2)^{-d/2}\exp\big\{-|x-m|^2/(2\sigma^2)\big\},\\
\rho_\theta(x)&=\sum_{i=1}^k e^{\alpha_i}\chi_\sigma(x;m_i).
\end{align}
For convenience in the writing, we will use the shortcut notation  $\chi_i(x):=\chi_\sigma(x;m_i)$ and $y_i(x):=x-m_i$. Differentiating with respect to the log-weights and locations gives
\begin{align}\label{eq:gaussian-tangent-weight}
d_i^\alpha(x)=e^{\alpha_i}\chi_i(x),\qquad d_{i,\ell}^m(x)=e^{\alpha_i}\chi_i(x)\frac{y_{i,\ell}(x)}{\sigma^2},
\end{align}
for $\ell=1,\ldots,d$. All event intensities are positive. We restrict to configurations where these directions are linearly independent; coinciding or redundant components can violate this condition.

\subsection{Gram matrix}\label{subsec:gaussian-gram-matrix}

\noindent The second ingredient is the Gram matrix $G(\theta)$ appearing in the projected system \eqref{eq:projected-ode}. Since its entries are the $L^2$ inner products of the tangent directions above, they can be computed explicitly from Gaussian overlaps. Define
\begin{align}\label{eq:gaussian-overlap}
K_{ij}(\theta)&=e^{\alpha_i+\alpha_j}(4\pi\sigma^2)^{-d/2}\exp\big\{-|m_i-m_j|^2/(4\sigma^2)\big\}\nonumber\\
&=\int_{\mathcal Z}e^{\alpha_i}\chi_i(x)e^{\alpha_j}\chi_j(x)\,dx.
\end{align}
Then, for $i,j=1,\ldots,k$ and $\ell,r=1,\ldots,d$,
\begin{align}
G_{\alpha_i,\alpha_j}&=K_{ij},\label{eq:gram-weight-weight}\\
G_{\alpha_j,m_{i,\ell}}&=G_{m_{i,\ell},\alpha_j}=-\frac{m_{i,\ell}-m_{j,\ell}}{2\sigma^2}K_{ij},\label{eq:gram-weight-location}\\
G_{m_{i,\ell},m_{j,r}}&=K_{ij}\bigg(\frac{\mathbbm 1_{\{\ell=r\}}}{2\sigma^2}-\frac{(m_{i,\ell}-m_{j,\ell})(m_{i,r}-m_{j,r})}{4\sigma^4}\bigg).\label{eq:gram-location-location}
\end{align}

\noindent To verify these identities, write $z=x-(m_i+m_j)/2$. Completion of the square gives
\[
\chi_i(x)\chi_j(x)=(2\pi\sigma^2)^{-d}\exp\bigg\{-\frac{|z|^2}{\sigma^2}-\frac{|m_i-m_j|^2}{4\sigma^2}\bigg\}.
\]
After normalization, $z$ has mean zero and covariance $(\sigma^2/2)I$. Also $y_i=z-(m_i-m_j)/2$ and $y_j=z+(m_i-m_j)/2$. The constant moment gives $G_{\alpha_i,\alpha_j}$, the mean of $y_i$ gives the mixed entry, and $\mathbb E[y_{i,\ell}y_{j,r}]=(\sigma^2/2)\mathbbm 1_{\{\ell=r\}}-(m_{i,\ell}-m_{j,\ell})(m_{i,r}-m_{j,r})/4$ gives the location block after division by $\sigma^4$.

\subsection{Projected generalized gradient}\label{subsec:gaussian-projected-force}

\noindent Having computed the Gram matrix $G(\theta)$, it remains to determine the projected generalized-gradient coefficients $g_T(\theta)$ appearing on the right-hand side of the projected system \eqref{eq:intro-projected-flow}. Direct differentiation of the Gaussian tangent directions gives
\begin{align}\label{eq:weight-divergence}
\nabla\cdot(\rho_\theta\nabla d_i^\alpha)(x)=e^{\alpha_i}\chi_i(x)\sum_{a=1}^k e^{\alpha_a}\chi_a(x)\bigg(\frac{y_a(x)\cdot y_i(x)+|y_i(x)|^2}{\sigma^4}-\frac d{\sigma^2}\bigg)
\end{align}
and
\begin{multline}\label{eq:location-divergence}
\nabla\cdot(\rho_\theta\nabla d_{i,\ell}^m)(x)\\
=e^{\alpha_i}\chi_i(x)\sum_{a=1}^k e^{\alpha_a}\chi_a(x)\bigg(\frac{y_{i,\ell}(x)\big(y_a(x)\cdot y_i(x)+|y_i(x)|^2\big)}{\sigma^6} -\frac{y_{a,\ell}(x)+(d+2)y_{i,\ell}(x)}{\sigma^4}\bigg).
\end{multline}
The corresponding mass pairings are obtained from the Gaussian overlap identities:
\begin{align}\label{eq:weight-mass-pairing}
\int_{\mathcal Z}\rho_\theta d_i^\alpha\,dx=\sum_{a=1}^kK_{ia},\qquad \int_{\mathcal Z}\rho_\theta d_{i,\ell}^m\,dx=-\frac1{2\sigma^2}\sum_{a=1}^k(m_{i,\ell}-m_{a,\ell})K_{ia}.
\end{align}
Substitution in \eqref{eq:projected-force-explicit} therefore gives
\begin{multline}\label{eq:gaussian-weight-force}
g_{T,\alpha_i}(\theta)=
-\frac{\kappa_W}{T}\sum_{q=1}^N\frac{\nabla\cdot(\rho_\theta\nabla d_i^\alpha)(x_q)}{\rho_\theta(x_q)+b_q}+\frac{\kappa_{FR}}{T}\sum_{q=1}^N\frac{\rho_\theta(x_q)d_i^\alpha(x_q)}{\rho_\theta(x_q)+b_q}
-\kappa_{FR}\sum_{a=1}^kK_{ia},
\end{multline}
and
\begin{align}\label{eq:gaussian-location-force}
g_{T,m_{i,\ell}}(\theta)={}&-\frac{\kappa_W}{T}\sum_{q=1}^N\frac{\nabla\cdot(\rho_\theta\nabla d_{i,\ell}^m)(x_q)}{\rho_\theta(x_q)+b_q}+\frac{\kappa_{FR}}{T}\sum_{q=1}^N\frac{\rho_\theta(x_q)d_{i,\ell}^m(x_q)}{\rho_\theta(x_q)+b_q}\nonumber\\
&+\frac{\kappa_{FR}}{2\sigma^2}\sum_{a=1}^k(m_{i,\ell}-m_{a,\ell})K_{ia}.
\end{align}

\subsection{Finite-dimensional implementation}\label{subsec:gaussian-implementation}

\noindent At each algorithmic time, the velocity is obtained by solving $G(\theta)v=g_T(\theta)$ rather than forming the inverse. For improved conditioning, the implementation may instead solve
\begin{align}\label{eq:gaussian-regularized-system}
\big(G(\theta)+\varepsilon_G I\big)v=g_T(\theta),\qquad \dot\theta=v,\qquad \varepsilon_G>0.
\end{align}
This changes the trajectory but not the exact equilibrium equations $g_T(\theta)=0$. The regularization is tied to the displayed parameter coordinates and does not remove geometric degeneracy of the original parametrization. The computation requires Gaussian evaluations, pairwise overlaps and a finite-dimensional linear solve. Initialization, time integration and numerical stopping criteria are specified in Section~\ref{sec:numerical-experiments}.

\section{Conditional consistency}\label{sec:consistency}

\noindent In this section we study whether equilibria of the projected dynamics on statistical manifolds, such as the Gaussian-mixture families introduced in Section~\ref{sec:gaussian-mixtures}, recover the true background intensity as more observations become available. We allow the chosen manifold to vary with the observation horizon, so that its tangent directions can become increasingly rich. Under the assumptions stated below, including that equilibria can be selected measurably on events whose probabilities tend to one, we prove convergence of these equilibrium densities to the true background.

\subsection{Heuristic considerations} Throughout the section, we assume that the observations available are generated by a Hawkes process with unknown true background density $\rho_\star$. Since consistency will be studied along increasingly rich statistical manifolds, the candidate densities themselves vary with the approximation scheme. We therefore place all densities under consideration, including the true background $\rho_\star$, inside a common ambient class $\mathcal K$. We will impose compactness and integrability assumptions on $\mathcal K$. With this ambient class fixed, we evaluate each candidate $\rho\in\mathcal K$ using the same observed catalog up to time $T$. This gives the random generalized gradient $\mathcal A_T(\rho)$ introduced in Section~\ref{sec:variational-formulation}. This operator prescribes the action used to define the projected dynamics at observation horizon $T$. The consistency argument proceeds in three fundamental steps.\\

\noindent First, we identify the deterministic limit of $\mathcal A_T(\rho)$ for a fixed candidate $\rho$. Since $\mathcal A_T(\rho)$ is built from time-averaged event contributions, an ergodic argument is the natural tool. The observed Hawkes process, however, is not stationary, as it starts at time zero with empty history. We therefore introduce a stationary Hawkes process with the same true background $\rho_\star$ and triggering mechanism, and define $\mathcal A_T^\circ(\rho)$ by replacing $(\eta,B_t)$ with its stationary counterpart $(\eta^\circ,B_t^\circ)$ while keeping $\rho$ fixed. The argument then has two separate parts. For the stationary process, the ergodic theorem gives convergence, when tested against each fixed $\phi\in H^\beta$, to the deterministic operator
\[
\mathcal A(\rho):=\mathbb E_\star^\circ[\mathcal A_1^\circ(\rho)].
\]
We then compare the stationary and empty-history processes and show that $\mathcal A_T(\rho)-\mathcal A_T^\circ(\rho)$ tends to zero in probability. It follows that the operator $\mathcal A_T(\rho)$ computed from the actual observations converges to the same deterministic limit $\mathcal A(\rho)$.\\

\noindent As part of this first step, for $\kappa_{FR}>0$, we also show that the limiting operator $\mathcal A$ identifies the true background, in the sense that
\[
\mathcal A(f)=0\quad\text{if and only if}\quad f=\rho_\star,\qquad f\in\mathcal K.
\]

\noindent Second, we strengthen this to convergence uniformly over $\rho\in\mathcal K$, so that the approximation remains valid when $\rho$ is replaced by the data-dependent equilibrium density appearing in the third step.\\

\noindent Finally, we combine the uniform convergence from the second step with a tangent-approximation condition imposed on the sequence of statistical manifolds, in order to show that the selected equilibrium densities approach $\rho_\star$. Let $\mathcal M_m$ be one of these manifolds and, whenever an equilibrium exists, let $\rho_{T,m}^\circ\in\mathcal M_m\cap\mathcal K$ denote a selected equilibrium density at observation horizon $T$. By definition, the density $\rho_{T,m}^\circ$ satisfies
\[
\langle\mathcal A_T(\rho_{T,m}^\circ),v\rangle=0
\qquad\text{for every }v\in T_{\rho_{T,m}^\circ}\mathcal M_m.
\]
In particular, $\mathcal A_T(\rho_{T,m}^\circ)$ is not required to vanish as a distribution on the whole test space; only its action on the tangent space of $\mathcal M_m$ at $\rho_{T,m}^\circ$ vanishes. The additional tangent-approximation assumption is precisely what allows these restricted equations to approximate the full test-space condition as the manifolds become richer. More precisely, if $v\in T_{\rho_{T,m}^\circ}\mathcal M_m$ approximates an admissible test function $\phi$, then the equilibrium condition gives
\[
\langle\mathcal A_T(\rho_{T,m}^\circ),\phi\rangle
=
\langle\mathcal A_T(\rho_{T,m}^\circ),\phi-v\rangle.
\]
Together with the uniform Sobolev bound proved below, tangent approximation makes the empirical residual small on the full test class; uniform convergence then transfers this to $\mathcal A(\rho_{T,m}^\circ)$, and compactness of $\mathcal K$ together with identification of the unique zero $\rho_\star$ yields consistency.

\subsection{Deterministic limit and identification}\label{subsec:consistency-deterministic-limit}

\noindent The first step is to identify the deterministic limit of $\mathcal A_T(\rho)$ for a fixed candidate $\rho$ and show that its zero equation identifies $\rho_\star$. We first construct the limit using stationary averages and compare it with the operator computed from the actual observations. We then determine which candidates make this limiting operator vanish. Throughout, we work on $\mathcal Z=\mathbb R^d$ under the subcritical ETAS conditions of Section~\ref{subsec:hawkes-etas}, with observations generated by the empty-history process of true intensity $\lambda_\star(t,x)=\rho_\star(x)+B_t(x)$.\\

\noindent Because this process is not stationary, a stationary ergodic theorem does not apply directly. We therefore introduce an auxiliary stationary Hawkes process with the same true background $\rho_\star$ and triggering mechanism, but with events extending into the past. Let $\boldsymbol\eta^\circ$ denote its marked counting measure, $\eta^\circ$ its ground process, and
\[
B_t^\circ(x)=\int_{(-\infty,t)}\int_{\mathcal Z\times\mathcal R}\nu(t-s,x-y,r)\,\boldsymbol\eta^\circ(ds,dy,dr)
\]
its excitation field. Its true intensity is $\lambda_\star^\circ(t,x)=\rho_\star(x)+B_t^\circ(x)$, and expectation under its stationary law is denoted by $\mathbb E_\star^\circ$. Proposition~\ref{prop:stationary-etas-regime} ensures that this process has finite mean event rate.\\

\noindent The following assumptions specify the common candidate class and control the averages used below. Fix $\beta>d/2+2$ and $M>0$, and let $\mathcal K$ be a nonempty subset of
\begin{align}\label{eq:admissible-density-class}
\big\{\rho\in H^\beta(\mathcal Z)\cap W^{1,1}(\mathcal Z):\ \|\rho\|_{H^\beta}+\|\rho\|_{W^{1,1}}\leq M,\quad\rho(x)>0\text{ for every }x\big\},
\end{align}
where $\|\rho\|_{W^{1,1}}=\int_{\mathcal Z}(|\rho|+|\nabla\rho|)\,dx$. We assume $\rho_\star\in\mathcal K$ and that $\mathcal K$ is compact for
\begin{align}\label{eq:compact-class-metric}
d_{\mathcal K}(\rho,\gamma)=\|\rho-\gamma\|_{H^\beta}+\|\rho-\gamma\|_{L^1}.
\end{align}
Compactness is imposed on $\mathcal K$ and does not follow from the norm bounds above. We further assume
\begin{align}\label{eq:localized-integrability}
\mathbb E_\star^\circ[Z]<\infty,\qquad Z:=\sup_{\rho\in\mathcal K}\int_{(0,1]\times\mathcal Z}\big(1+|\nabla\log\rho(x)|\big)\,\eta^\circ(dt,dx).
\end{align}
The supremum in \eqref{eq:localized-integrability} is taken after integration over the unit event block. The logarithmic derivative appears because, for $b\geq0$, the complete empirical integrand satisfies
\begin{align}\label{eq:force-factorization}
&\frac{-\kappa_W\nabla\cdot(\rho\nabla\phi)(x)+\kappa_{FR}\rho(x)\phi(x)}{\rho(x)+b}\nonumber\\
&\qquad=\frac{\rho(x)}{\rho(x)+b}\big[-\kappa_W\Delta\phi(x)-\kappa_W\nabla\log\rho(x)\cdot\nabla\phi(x)+\kappa_{FR}\phi(x)\big].
\end{align}
Since $\rho/(\rho+b)\leq1$, the absolute value of \eqref{eq:force-factorization} is bounded by $C_0(1+|\nabla\log\rho(x)|)\|\phi\|_{H^\beta}$, with $C_0:=C_\beta(\kappa_W+\kappa_{FR})$. Thus \eqref{eq:localized-integrability} controls the event contributions even when $\rho$ becomes small. On each fixed ball $K_R:=\{x:|x|\leq R\}$, compactness and strict positivity additionally give a uniform positive lower bound for $\rho$. No global reciprocal bound or common upper and lower density profile is assumed.\\

\noindent For $\rho\in\mathcal K$, replace $(\eta,B_t)$ by $(\eta^\circ,B_t^\circ)$ in \eqref{eq:empirical-generalized-gradient-integral}, without changing the candidate density. This defines the auxiliary random operator, for $T>0$ and $\phi\in H^\beta$,
\begin{align}\label{eq:consistency-stationary-operator}
\langle\mathcal A_T^\circ(\rho),\phi\rangle
={}&\frac1T\int_{(0,T]\times\mathcal Z}
\frac{-\kappa_W\nabla\cdot(\rho\nabla\phi)(x)+\kappa_{FR}\rho(x)\phi(x)}
{\rho(x)+B_t^\circ(x)}
\,\eta^\circ(dt,dx)\nonumber\\
&-\kappa_{FR}\int_{\mathcal Z}\rho(x)\phi(x)\,dx.
\end{align}
We obtain a deterministic operator by taking the expectation of the stationary expression over one unit of time:
\begin{align}\label{eq:consistency-stationary-mean}
\langle\mathcal A(\rho),\phi\rangle
:=
\mathbb E_\star^\circ\big[\langle\mathcal A_1^\circ(\rho),\phi\rangle\big].
\end{align}
This expectation is finite by \eqref{eq:force-factorization} and \eqref{eq:localized-integrability}. Stationarity and normalization by $T$ imply that averaging $\mathcal A_T^\circ(\rho)$ gives the same operator for every $T>0$.\\

\noindent We next show why this expectation is the limit of the empirical operator. For each fixed $\rho\in\mathcal K$ and $\phi\in H^\beta$, the integrability assumption permits application of the ergodic theorem to the stationary unit-block contributions. The final partial block is controlled by the same integrable block bound, as detailed in Appendix~\ref{app:uniform-convergence}. Thus, as $T$ increases without bound,
\[
\langle\mathcal A_T^\circ(\rho),\phi\rangle
\stackrel{\mathrm{a.s.}}{\rightarrow}
\langle\mathcal A(\rho),\phi\rangle.
\]
The coupling comparison in Appendix~\ref{app:uniform-convergence} shows that $\langle\mathcal A_T(\rho)-\mathcal A_T^\circ(\rho),\phi\rangle$ tends to zero in probability. Consequently, $\langle\mathcal A_T(\rho),\phi\rangle\stackrel{\mathbb P}{\rightarrow}\langle\mathcal A(\rho),\phi\rangle$, so the operator computed from the actual observations has the same deterministic limit.\\

\noindent To identify the zeros of $\mathcal A$, we compute its action using the stationary compensator. Finite stationary mean rate implies that $B_0^\circ(x)$ is finite almost surely for almost every $x$. At those points, define
\begin{align}\label{eq:population-score-consistency}
\mathfrak s_\rho(x)
&=
\mathbb E_\star^\circ\bigg[\frac{\lambda_\star^\circ(0,x)}{\rho(x)+B_0^\circ(x)}\bigg]-1
=
\big(\rho_\star(x)-\rho(x)\big)\mathbb E_\star^\circ\bigg[\frac1{\rho(x)+B_0^\circ(x)}\bigg].
\end{align}
The expected ratio is finite because it is bounded by $\max\{1,\rho_\star(x)/\rho(x)\}$. Using the compensator identity and $\int_{\mathcal Z}\nabla\cdot(\rho\nabla\phi)\,dx=0$, justified by $\rho\in W^{1,1}$ and Sobolev embedding, gives
\begin{align}\label{eq:limiting-generalized-gradient}
\langle\mathcal A(\rho),\phi\rangle
=
\int_{\mathcal Z}\mathfrak s_\rho(x)\big[-\kappa_W\nabla\cdot(\rho\nabla\phi)(x)+\kappa_{FR}\rho(x)\phi(x)\big]\,dx.
\end{align}
The integral is well defined because
\[
\rho|\mathfrak s_\rho|\leq|\rho_\star-\rho|,
\qquad
|\nabla\rho|\,|\mathfrak s_\rho|
\leq\rho_\star|\nabla\log\rho|+|\nabla\rho|,
\]
and these quantities are controlled by the class assumptions and \eqref{eq:localized-integrability}.\\

\noindent The following result identifies the only zero of $\mathcal A$ and gives the continuity needed to deduce consistency from a small limiting residual.

\begin{proposition}\label{prop:population-identification}
Under the class, model and integrability assumptions above, $\mathcal A(\rho)\in H^{-\beta}$ for every $\rho\in\mathcal K$, and $\rho\mapsto\mathcal A(\rho)$ is continuous from $(\mathcal K,d_{\mathcal K})$ into $H^{-\beta}$. If $\kappa_{FR}>0$, then $\mathcal A(\rho)=0$ if and only if $\rho=\rho_\star$.
\end{proposition}

\noindent Appendix~\ref{app:localization} proves continuity by dominated convergence and identification by a cutoff argument when $\kappa_W>0$; when $\kappa_W=0$, identification follows directly from the reaction term.

\subsection{Uniform convergence}\label{subsec:consistency-uniform-convergence}

\noindent The second step makes the preceding approximation valid at data-dependent equilibrium densities. We strengthen convergence for each fixed candidate and test function to a bound uniform over $\rho\in\mathcal K$ and a test-function unit ball. To specify that ball and the corresponding distributional norm, set, for $\gamma>0$,
\[
\mathcal X^\gamma:=\{\phi\in H^\gamma:\ |x|\phi\in L^2\},\qquad\|\phi\|_{\mathcal X^\gamma}:=\|\phi\|_{H^\gamma}+\||x|\phi\|_{L^2},
\]
and denote its continuous dual by $\mathcal X^{-\gamma}$. In particular, for $F\in H^{-\beta}$,
\[
\|F\|_{\mathcal X^{-(\beta+r)}}=\sup_{\|\phi\|_{\mathcal X^{\beta+r}}\leq1}|\langle F,\phi\rangle|.
\]
For every $r>0$, $\mathcal X^{\beta+r}$ is dense in $H^\beta$ and its unit ball is relatively compact in $H^\beta$. The additional derivatives provide local compactness, while the spatial weight controls the tails. Subsection~\ref{appsubsec:test-functions} proves this compact inclusion.

\noindent The following result bounds the approximation error uniformly over $\rho\in\mathcal K$ and this test-function unit ball. The accompanying $H^{-\beta}$ bound controls the tangent approximation error in the third step.

\begin{proposition}\label{prop:uniform-generalized-gradient-convergence}
Under the same class, model and integrability assumptions, for every $r>0$,
\begin{align}\label{eq:uniform-generalized-gradient-convergence}
\sup_{\rho\in\mathcal K}\|\mathcal A_T(\rho)-\mathcal A(\rho)\|_{\mathcal X^{-(\beta+r)}}\stackrel{\mathbb P}{\rightarrow}0
\end{align}
as $T$ increases without bound, and
\begin{align}\label{eq:uniform-empirical-gradient-bound}
\sup_{\rho\in\mathcal K}\|\mathcal A_T(\rho)\|_{H^{-\beta}}=O_{\mathbb P}(1).
\end{align}
\end{proposition}

\noindent Appendix~\ref{app:uniform-convergence} proves these estimates using the full stationary event blocks, with spatial localization only in the comparison with the empty-history process.

\subsection{Tangent approximation and consistency}\label{subsec:consistency-tangent-approximation}

\noindent The third step combines uniform convergence with an additional assumption on the tangent spaces. An equilibrium makes the empirical operator vanish only on the tangent space at its density. To control its action on the test functions used above, let $(\mathcal M_m)_{m\geq1}$ be a sequence of statistical manifolds as in Subsection~\ref{subsec:projected-equations} and impose the following approximation condition: for some $r>0$ and nonnegative numbers $\varepsilon_m$ with $\lim_{m\to\infty}\varepsilon_m=0$, every $\rho\in\mathcal M_m\cap\mathcal K$ and $\phi\in\mathcal X^{\beta+r}$ admit $v\in T_\rho\mathcal M_m$ satisfying
\begin{align}\label{eq:tangent-approximation-consistency}
\|\phi-v\|_{H^\beta}\leq\varepsilon_m\|\phi\|_{\mathcal X^{\beta+r}}.
\end{align}
This is a uniform approximation assumption on the tangent spaces at the candidate densities. Compactness of the test-function unit ball allows approximation by some finite-dimensional spaces, but does not establish \eqref{eq:tangent-approximation-consistency} for the tangent spaces of the chosen manifolds.

\begin{theorem}\label{thm:projected-consistency}
Assume the class, model and integrability conditions above, $\kappa_{FR}>0$, and \eqref{eq:tangent-approximation-consistency}. Let $\lim_{n\to\infty}T_n=\infty$ and $\lim_{n\to\infty}\varepsilon_{m_n}=0$. For every $n$, let $\rho_n^\circ$ be a measurable $\mathcal K$-valued estimator and $E_n$ a measurable event, both determined by observations up to $T_n$. Suppose $\lim_{n\to\infty}\mathbb P[E_n]=1$ and that, on $E_n$, $\rho_n^\circ\in\mathcal M_{m_n}\cap\mathcal K$ and
\begin{align}\label{eq:empirical-projected-equilibrium-consistency}
\langle\mathcal A_{T_n}(\rho_n^\circ),v\rangle=0\qquad\text{for every }v\in T_{\rho_n^\circ}\mathcal M_{m_n}.
\end{align}
Then
\begin{align}\label{eq:projected-consistency-conclusion}
d_{\mathcal K}(\rho_n^\circ,\rho_\star)\stackrel{\mathbb P}{\rightarrow}0.
\end{align}
In particular, the densities converge in probability in both $H^\beta(\mathcal Z)$ and $L^1(\mathcal Z)$, and $\int_{\mathcal Z}\rho_n^\circ\,dx$ converges in probability to $\int_{\mathcal Z}\rho_\star\,dx$.
\end{theorem}

\begin{proof}
We first show that the limiting generalized gradient vanishes asymptotically at the selected density. Fix $n$ and work on $E_n$. For $\phi\in\mathcal X^{\beta+r}$, choose $v_n\in T_{\rho_n^\circ}\mathcal M_{m_n}$ satisfying \eqref{eq:tangent-approximation-consistency}. By the equilibrium equation,
\begin{align*}
\langle\mathcal A(\rho_n^\circ),\phi\rangle
={}&\langle\mathcal A(\rho_n^\circ)-\mathcal A_{T_n}(\rho_n^\circ),\phi\rangle+\langle\mathcal A_{T_n}(\rho_n^\circ),\phi-v_n\rangle.
\end{align*}
The first term measures the difference between the empirical and limiting operators; the second is the error from replacing $\phi$ by a tangent direction. Applying the corresponding norm bounds and taking the supremum over $\|\phi\|_{\mathcal X^{\beta+r}}\leq1$ gives, on $E_n$,
\begin{align}\label{eq:limiting-gradient-at-empirical-equilibrium}
\|\mathcal A(\rho_n^\circ)\|_{\mathcal X^{-(\beta+r)}}\leq{}&\sup_{\rho\in\mathcal K}\|\mathcal A_{T_n}(\rho)-\mathcal A(\rho)\|_{\mathcal X^{-(\beta+r)}}\nonumber\\
&+\varepsilon_{m_n}\sup_{\rho\in\mathcal K}\|\mathcal A_{T_n}(\rho)\|_{H^{-\beta}}.
\end{align}
The tangent approximant is used only to obtain this pointwise inequality, so no measurable selection of $v_n$ is needed. Both terms on the right tend to zero in probability by Proposition~\ref{prop:uniform-generalized-gradient-convergence} and $\lim_{n\to\infty}\varepsilon_{m_n}=0$. For every $a>0$,
\begin{align*}
\mathbb P\big[\|\mathcal A(\rho_n^\circ)\|_{\mathcal X^{-(\beta+r)}}>a\big]
\leq{}&\mathbb P[E_n^c]+\mathbb P\big[E_n\cap\{\|\mathcal A(\rho_n^\circ)\|_{\mathcal X^{-(\beta+r)}}>a\}\big].
\end{align*}
The first probability tends to zero by assumption, and the second by \eqref{eq:limiting-gradient-at-empirical-equilibrium}. Hence
\begin{align}\label{eq:population-residual-zero}
\|\mathcal A(\rho_n^\circ)\|_{\mathcal X^{-(\beta+r)}}\stackrel{\mathbb P}{\rightarrow}0.
\end{align}

\noindent To pass from residual convergence to density convergence, fix $\delta>0$ and set $\mathcal K_\delta:=\{\rho\in\mathcal K:d_{\mathcal K}(\rho,\rho_\star)\geq\delta\}$. If this set is empty, the conclusion is immediate. Otherwise, it is compact for $d_{\mathcal K}$. Proposition~\ref{prop:population-identification} and the continuous inclusion $\mathcal X^{\beta+r}\subset H^\beta$ make $\rho\mapsto\|\mathcal A(\rho)\|_{\mathcal X^{-(\beta+r)}}$ continuous. This norm is strictly positive on $\mathcal K_\delta$: if it vanished, density of $\mathcal X^{\beta+r}$ in $H^\beta$ would give $\mathcal A(\rho)=0$ and hence $\rho=\rho_\star$. Therefore
\[
c_\delta:=\min_{\rho\in\mathcal K_\delta}\|\mathcal A(\rho)\|_{\mathcal X^{-(\beta+r)}}>0.
\]
Since $\rho_n^\circ$ is $\mathcal K$-valued on all samples,
\[
\mathbb P\big[d_{\mathcal K}(\rho_n^\circ,\rho_\star)\geq\delta\big]\leq\mathbb P\big[\|\mathcal A(\rho_n^\circ)\|_{\mathcal X^{-(\beta+r)}}\geq c_\delta\big].
\]
The right-hand side tends to zero by \eqref{eq:population-residual-zero}, proving convergence in $d_{\mathcal K}$. The two norm conclusions follow from \eqref{eq:compact-class-metric}, and convergence of the total mass follows from $|\int_{\mathcal Z}(\rho_n^\circ-\rho_\star)\,dx|\leq\|\rho_n^\circ-\rho_\star\|_{L^1}$.
\end{proof}

\noindent The theorem is conditional on the tangent-approximation property and on the high-probability existence of measurable empirical equilibria. These two conditions are not established here for the Gaussian-mixture implementation of Section~\ref{sec:gaussian-mixtures}. Accordingly, Theorem~\ref{thm:projected-consistency} applies to projected estimating schemes satisfying the stated assumptions, rather than establishing consistency of that particular implementation. Outside $E_n$, the estimator may be assigned any fixed member of $\mathcal K$. The conclusion concerns the selected equilibrium densities, not convergence of projected trajectories or likelihood maximization.

\subsection{Gaussian compatibility}\label{subsec:consistency-gaussian-classes}

\noindent The three-step consistency argument is now complete. We finish by checking which of its assumptions are verified by the Gaussian classes of Section~\ref{sec:gaussian-mixtures}. The next proposition supplies a compact ambient class containing mixtures with any finite number of components, provided their bandwidth is fixed, their centers remain in a fixed ball, and their total mass stays in a fixed positive interval. A finite first spatial moment of the stationary process then suffices for the event-integrability assumption.

\begin{proposition}\label{prop:gaussian-localization}
Fix $\sigma>0$, $0\leq D<\infty$ and $0<a\leq b<\infty$. The Gaussian convolutions
\[
\rho(x)=\int_{\{|m|\leq D\}}\chi_\sigma(x;m)\,\xi(dm),\qquad a\leq\xi[\mathbb R^d]\leq b,
\]
with finite positive mixing measures supported on $\{|m|\leq D\}$ form a compact class for $d_{\mathcal K}$ satisfying \eqref{eq:admissible-density-class} for some $M$. Moreover,
\begin{align}\label{eq:gaussian-log-derivative-bound}
|\nabla\log\rho(x)|\leq\frac{|x|+D}{\sigma^2}.
\end{align}
Thus \eqref{eq:localized-integrability} holds whenever
\begin{align}\label{eq:stationary-spatial-first-moment}
\mathbb E_\star^\circ\bigg[\int_{(0,1]\times\mathcal Z}(1+|x|)\,\eta^\circ(dt,dx)\bigg]<\infty.
\end{align}
\end{proposition}

\noindent Appendix~\ref{app:gaussian-class} proves the proposition and gives sufficient model conditions for \eqref{eq:stationary-spatial-first-moment}. The logarithmic-derivative bound is independent of the number of components and requires no lower bound on individual component weights. This verifies the analytical class conditions, not all the hypotheses of Theorem~\ref{thm:projected-consistency}: membership of $\rho_\star$, tangent approximation and high-probability existence of equilibria must still be checked for the chosen scheme. In particular, increasing the number of Gaussian components alone does not imply \eqref{eq:tangent-approximation-consistency}; this condition must be verified uniformly over the candidate configurations in the chosen manifolds. Fixed-bandwidth mixtures are not asserted to approximate arbitrary true densities.


\section{Numerical experiments}\label{sec:numerical-experiments}

Next, we present numerical applications of the intenstiy estimation scheme developed above. Beyond the direct implementation of the theoretical scheme, its numerical realization brings additional phenomena that must be taken into account. We therefore begin by discussing a few considerations arising at this level. We then illustrate the methodology in two complementary settings: (i) a simulated non-homogeneous Poisson process and (ii) a real-data ETAS application to seismicity in Southern Mexico.\\

\noindent In Appendix~\ref{sec:pseudocode}, we present the pseudocode for the algorithm proposed in Section~\ref{sec:gaussian-mixtures}. Furthermore, for readers interested in replicating this work, the codes used to generate all examples are available in the GitHub repository \href{https://github.com/isaiasmanuel/VariationalHawkes}{isaiasmanuel/VariationalHawkes}.\\

\subsection{Numerical considerations}

\noindent For the computational implementation, rather than directly using standard gradient descent, we have chosen to incorporate the ADAM algorithm \cite{kingma2014adam}. As noted by authors such as \cite{corrales2025annealed}, incorporating this algorithm in the context of particle-based variational inference accelerates convergence. Furthermore, this approach avoids numerical issues arising from differing orders of magnitude between the gradients with respect to the weights and the means.\\

\noindent In our numerical experiments, we observed a common behavior where different centers $\mathbf{m}_i$ tended to collapse to the same value. When this occurred, the Gram matrix became non-invertible, leading to numerical instability in the implementation. To prevent this issue, we introduced a merging threshold (line 17 of the pseudocode in Appendix~\ref{sec:pseudocode}). Under this threshold, whenever two mixture components become sufficiently close, they are merged into a single component located at their average position, with its weight set to the sum of their individual weights.\\

\noindent Far from being a drawback, this collapsing behavior is actually advantageous. This automatically reduces the model's complexity without requiring model selection procedures, such as  Akaike information criterion (AIC) \cite{akaike1987factor}, cross-validation, or Reversible Jump MCMC \cite{green1995reversible}.\\

\noindent Although $\kappa_W$ and $\kappa_{FR}$ are introduced as algorithm hyperparameters in Equation~\eqref{eq:intro-generalized-gradient}, having no prior preference for either metric, we set $\kappa_W = \kappa_{FR} = 1$ across all experiments in this work. The variance of the mixture components, $\sigma^2$, will be the only hyperparameter to be calibrated. To this end, for each example, we evaluated the model over a grid of $\sigma^2$ values and selected the optimal model using AIC or Bayesian Information Criterion (BIC) \cite{schwarz1978estimating}. Nevertheless,  the optimal models selected by AIC and BIC in  our experiments were the same, therefore we omit further discussion regarding BIC to streamline the text.\\

\subsection{Simulated Poisson Process}

While the majority of this work focuses on Hawkes processes, inference for Poisson processes remains a subject of active interest \cite{xiao2015modeling,banales2026graphical}; for this reason, we include an example of how to estimate its intensity. Also, in the context of Hawkes process inference, \cite{veen2008estimation} suggested using the Expectation-Maximization algorithm \cite{dempster1977maximum} to estimate the intensity function, which requires maximizing a likelihood function that depends solely on a Poisson process.

To evaluate the performance of the algorithm proposed in Section~\ref{sec:gaussian-mixtures}, we define a Poisson process with the following intensity function:

\begin{equation}
\label{Eq:IntensidadPoisson}
\lambda(\mathbf{x}) = 70 \left(\frac{1}{2} \Big( \frac{2}{3} \chi_\sigma(\mathbf{x}; \mathbf{m}_1) + \frac{1}{3} \chi_\sigma(\mathbf{x}; \mathbf{m}_2) \Big) + \frac{1}{2} \Big( \frac{1}{3} \chi_\sigma(\mathbf{x}; \mathbf{m}_3) + \frac{2}{3} \chi_\sigma(\mathbf{x}; \mathbf{m}_4) \Big)\right),
\end{equation}

where $\mathbf{m}_1 = (0, 0)$, $\mathbf{m}_2 = (2, 2)$, $\mathbf{m}_3 = (4, 6)$, $\mathbf{m}_4 = (6, 2)$, and $\sigma = 1$. And we perform a stochastic simulation of this process using the thinning algorithm \cite{lewis1979simulation,illian2008statistical} to obtain a sample of 695 points, which can be reproduced using \texttt{Simulation.py} from our GitHub repository.

\begin{table}[htbp]

\centering
\caption{Model evaluation metrics over a grid search of $\sigma^2$ bandwidth values for the Poisson process case.}
\label{Tab:EjemploPoisson}
\begin{tabular}{cccccc}
$\sigma^2$ & AIC & BIC & Log-Likelihood & Components ($k$) & Parameters \\
0.10 & 171498.49 & 172057.39 & -85626.24 & 41 & 123 \\
0.35 & 3246.64   & 3587.44   & -1548.32  & 25 & 75  \\
0.60 & 1471.30   & 1703.04   & -684.65   & 17 & 51  \\
0.85 & 1260.48   & 1437.70   & -591.24   & 13 & 39  \\
\textbf{1.10} & \textbf{1194.08} & \textbf{1303.13} & \textbf{-573.04} & \textbf{8} & \textbf{24} \\
1.35 & 1227.60   & 1309.39   & -595.80   & 6  & 18  \\
1.60 & 1292.70   & 1374.49   & -628.35   & 6  & 18  \\
1.85 & 1347.01   & 1401.54   & -661.50   & 4  & 12  \\
2.10 & 1427.34   & 1522.77   & -692.67   & 7  & 21  \\
2.35 & 1484.81   & 1580.23   & -721.41   & 7  & 21  \\
2.60 & 1519.21   & 1573.74   & -747.60   & 4  & 12  \\
2.85 & 1566.70   & 1621.23   & -771.35   & 4  & 12  \\
3.10 & 1616.51   & 1671.04   & -796.26   & 4  & 12  \\
3.35 & 1676.67   & 1758.46   & -820.33   & 6  & 18  \\
3.60 & 1703.94   & 1744.84   & -842.97   & 3  & 9   \\
3.85 & 1770.21   & 1865.63   & -864.11   & 7  & 21  \\
4.10 & 1785.46   & 1826.36   & -883.73   & 3  & 9   \\
4.35 & 1845.77   & 1941.19   & -901.88   & 7  & 21  \\
4.60 & 1867.22   & 1935.38   & -918.61   & 5  & 15  \\
4.85 & 1903.95   & 1985.74   & -933.98   & 6  & 18  \\
5.10 & 1914.11   & 1955.00   & -948.05   & 3  & 9   \\

\end{tabular}
\end{table}

Table~\ref{Tab:EjemploPoisson} summarizes the AIC and BIC values obtained across the candidate values of $\sigma^2$ when initializing the algorithm with $k=100$ components. Based on these metrics, we conclude that the optimal model is achieved at $\sigma^2=1.1$.\\ 

\begin{figure}[htbp]
    \centering
    \begin{subfigure}[b]{0.48\linewidth}
        \centering
        \includegraphics[width=\linewidth]{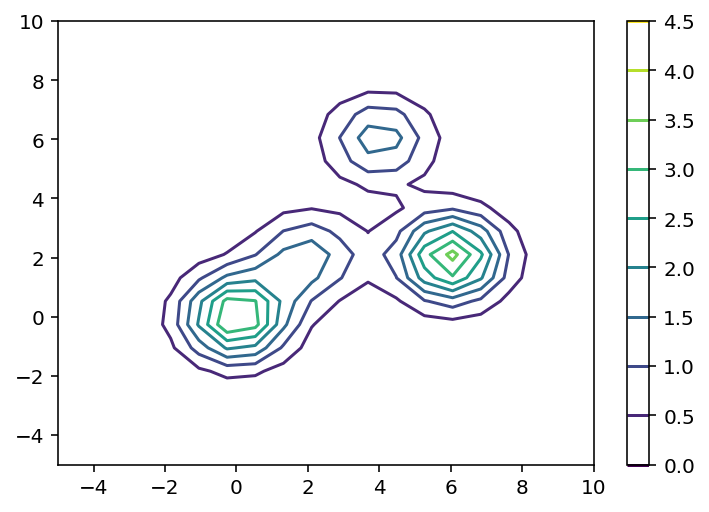} 
        \caption{True intensity.}
        \label{fig:real}
    \end{subfigure}
    \hfill 
    \begin{subfigure}[b]{0.48\linewidth}
        \centering
        \includegraphics[width=\linewidth]{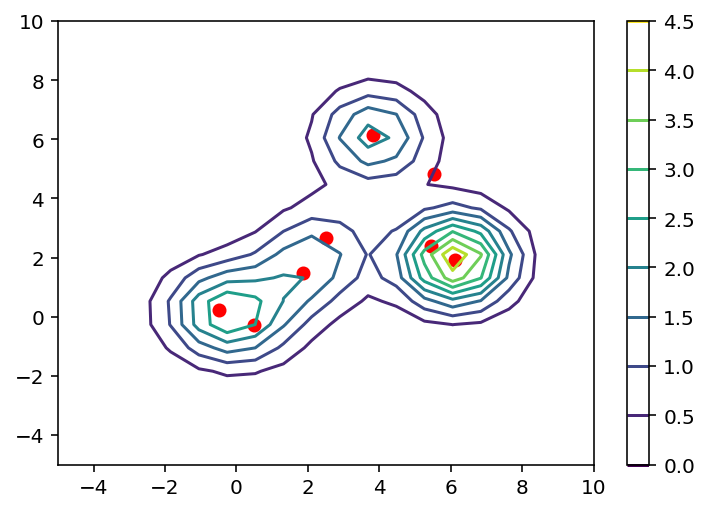} 
        \caption{Best model intensity ($\sigma^2 = 1.1$).}
        \label{fig:estimada}
    \end{subfigure}
    \caption{Comparison of the process intensity function. The red dots represent the $m_i$ values of the mixture elements}
    \label{fig:comparacion_poisson}
\end{figure}

\noindent Finally, Figure~\ref{fig:comparacion_poisson} compares the true and estimated intensity functions for the Poisson process. The left panel displays the contour lines of the ground-truth intensity function defined in \eqref{Eq:IntensidadPoisson}, while the right panel shows the intensity function estimated by our algorithm, which yields a highly satisfactory approximation. Notably, evaluating our estimated intensity requires computing only 8 Gaussian components, in stark contrast to the 695 evaluations that standard Kernel Density Estimation (KDE) would require. Generating the grid search results in Table~\ref{Tab:EjemploPoisson} required 166.24 seconds of computation time on a MacBook Pro equipped with an M5 processor.

\subsection{Seismicity data in southern Mexico}
As discussed in Section~\ref{sec:Existing}, several methods exist for estimating intensity functions. A prominent application of Hawkes process inference lies in seismology within the framework of Epidemic Type Aftershock Sequence (ETAS) models, where recent literature has actively explored southern Mexico \cite{banales2025inhomogeneous,jaimes2025aftershock,banales2026graphical}. To demonstrate the capability of our algorithm, this section presents ETAS process inference applied to seismic data from this region.\\

\noindent For this analysis, we utilize the U.S. Geological Survey (USGS) catalog covering the period from January 1, 2000, to December 31, 2016, within the spatial region bounded by $14.460^\circ\text{N}$--$20.475^\circ\text{N}$ latitude and $106.061^\circ\text{W}$--$95.412^\circ\text{W}$ longitude, which has a cutoff magnitude of 4.3. For a more detailed discussion regarding the dataset selection, we refer the reader to \cite{banales2025inhomogeneous}.

We define the conditional intensity function according to the space-time ETAS model presented in \cite{zhuang2002stochastic}, i.e.,

\begin{equation}
    \lambda(t, x, y \mid \mathcal{H}_t) = \mu(x, y) + \sum_{\{k : t_k < t\}} k(M_k) g(t - t_k) f(x - x_k, y - y_k \mid M_k),
\end{equation}
the index $k \in \{1, \dots, n\}$ denotes the $k$-th event, characterized by its spatial coordinates $(x_k, y_k)$, occurrence time $t_k$, and magnitude $M_k$. Without loss of generality, the arrival times are assumed to be strictly ordered chronologically ($t_1 < t_2 < \dots < t_n$). In addition,

\begin{equation}
    k(M) = A e^{\alpha(M - M_0)}
\end{equation}

\begin{equation}
    g(t) = (p - 1) c^{p-1} (t + c)^{-p} \mathbb{1}(t > 0)
\end{equation}

\begin{equation}
    f(x, y \mid M) = \frac{1}{2\pi \sqrt{d e^{\alpha(M - M_0)}}} \exp \left\{ -\frac{1}{2} \frac{x^2+y^2}{de^{\alpha(M - M_0)}} \right\}.
\end{equation}
Thus, within this framework, the objective is to estimate the spatially varying background intensity $\mu(x,y)$ alongside the ETAS parameters $(A, \alpha, c, p, d)$.\\

\noindent Following a procedure similar to that of the Poisson baseline, Table~\ref{Tab:ETAS_GridSearch} summarizes the AIC and BIC metrics across the candidate models. In this case, the optimal fit is achieved at $\sigma^2 = 0.6$, comprising only 6 mixture components. The full grid search across $\sigma^2$ values presented in Table~\ref{Tab:ETAS_GridSearch} required a total execution time of $734.79$ seconds on the same workstation used for the Poisson process benchmark.\\

\noindent The estimated self-exciting parameter vector $(A, \alpha, c, p, d)$ yields (0.18\text{ events}, 0.97, 0.02\text{ days}, 1.26, 0.0065$^\circ)$, which closely aligns with the values reported by \cite{banales2025inhomogeneous} for the same region using a modified methodology from \cite{li2020space}, namely $(0.11\text{ events}, 1.11, 0.02\text{ days}, 1.36, 0.0048^\circ)$. For comparison, we applied Zhuang's ETAS model \cite{zhuang2002stochastic} to this same catalog, obtaining parameter values of $(A, \alpha, c, p, d) = (0.679124, 0.926814, 0.009243\text{ days}, 1.039849, 0.007568^\circ)$.\\

\noindent The optimum based on \cite{zhuang2002stochastic} is a supercritical point process \cite{helmstetter2003mainshocks} similar to those reported by \cite{jaimes2025aftershock} for the same region, which means that the process may exhibit explosive behavior. Consequently, it is not physically preferable to the optimum achieved using our methodology, as subcritical regimes guarantee that the aftershock cascade remains finite and stable, with the event frequency progressively decaying over time until returning to the background seismicity rate.\\

\noindent A key distinction between our methodology and those proposed by \cite{zhuang2002stochastic} and \cite{schoenberg2013facilitated} is that our algorithm does not estimate the branching structure at each iteration; hence, it does not utilize latent variable information and relies solely on coordinate-wise optimization via profile likelihoods. In contrast, both benchmark approaches require evaluating the background rate $\mu$ at every event location during each update step, which, in the case of \cite{zhuang2002stochastic}, entails evaluating as many Gaussian components as there are data points in the catalog.\\

\begin{table}[htbp]
\centering
\caption{Model evaluation metrics over a grid search of $\sigma^2$ bandwidth values for the Hawkes process case.}
\label{Tab:ETAS_GridSearch}
\begin{tabular}{cccccc}
$\sigma^2$ & AIC & BIC & Log-Likelihood & Components ($k$) & Parameters \\
0.10 & 30448.32 & 30626.05 & -15186.16 & 11 & 38 \\
0.20 & 10954.69 & 11076.30 & -5451.35 & 7 & 26 \\
0.30 & 8497.65 & 8619.26 & -4222.83 & 7 & 26 \\
0.40 & 7957.68 & 8065.26 & -3955.84 & 6 & 23 \\
0.50 & 7852.39 & 7959.96 & -3903.19 & 6 & 23 \\
\textbf{0.60} & \textbf{7851.05} & \textbf{7958.63} & \textbf{-3902.53} & \textbf{6} & \textbf{23} \\
0.70 & 7880.11 & 7973.66 & -3920.06 & 5 & 20 \\
0.80 & 7921.70 & 8015.24 & -3940.85 & 5 & 20 \\
0.90 & 7960.26 & 8039.77 & -3963.13 & 4 & 17 \\
1.00 & 8006.53 & 8086.04 & -3986.26 & 4 & 17 \\
\end{tabular}
\end{table}

\begin{figure}[htbp]
    \centering
    \begin{subfigure}[b]{0.48\linewidth}
        \centering
        \includegraphics[width=\linewidth]{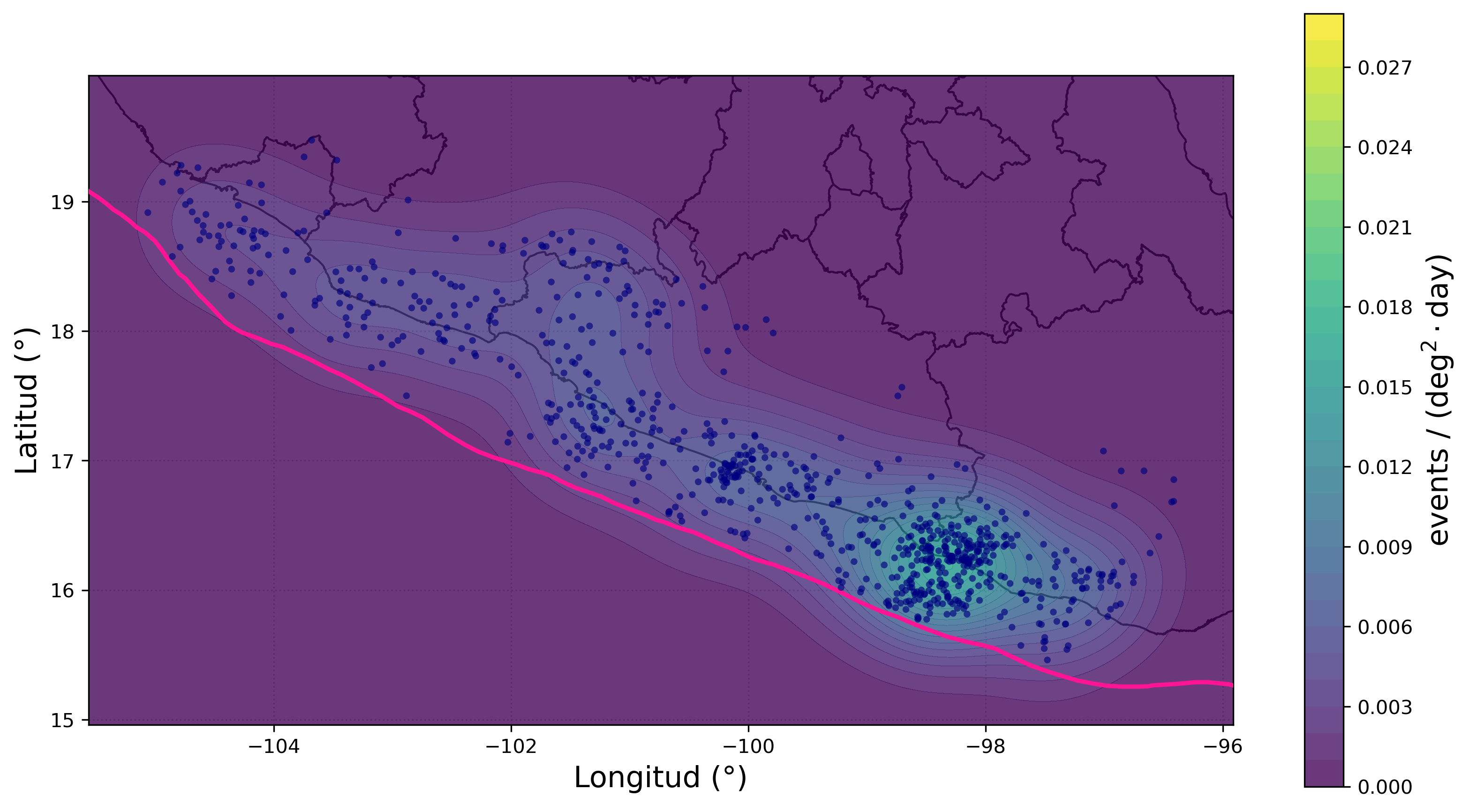} 
        \caption{Using \cite{zhuang2002stochastic} methodology.}
        \label{fig:MapEtas}
    \end{subfigure}
    \hfill 
    \begin{subfigure}[b]{0.48\linewidth}
        \centering
        \includegraphics[width=\linewidth]{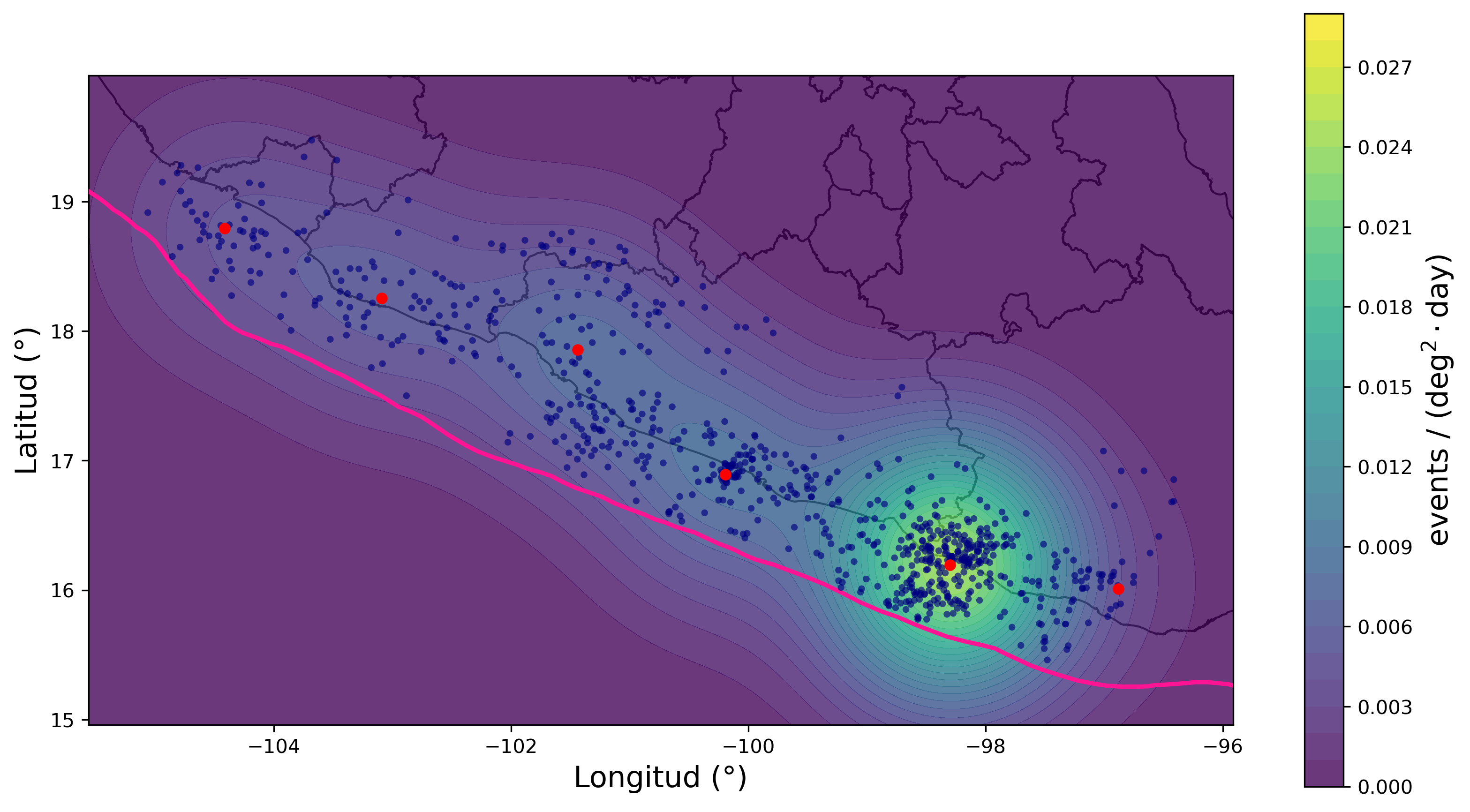} 
        \caption{Best model intensity ($\sigma^2 = 0.6$).}
        \label{fig:Mapa}
    \end{subfigure}
    \caption{Spatial distribution of seismic events in Southern Mexico from the USGS catalog (2000–2016) and the estimated ETAS background intensity function. The Middle America Trench is shown in pink, and the red dots represent the mixture element centers $m_i$, ($i=1,2,…,6$).}
    \label{fig:mapa_etas}
\end{figure}

\noindent Figure~\ref{fig:Mapa} presents the estimated background rate $\mu$ from the best ETAS model in Table~\ref{Tab:ETAS_GridSearch}, which is consistent with those reported for the region \cite{banales2025inhomogeneous,banales2026graphical} and with the map obtained using the methodology of \cite{zhuang2002stochastic} presented in Figure \ref{fig:MapEtas}. This agreement allows us to conclude that our methodology is effective for estimating the ETAS process.\\

\noindent Although our algorithm does not offer an advantage in terms of computational time, taking $71\text{ seconds}$ to fit the best model compared to $16.79\text{ seconds}$ using \cite{zhuang2002stochastic}, we believe it represents a valuable proposal, as it achieves a parsimonious model and a physically meaningful optimum without requiring the inclusion of latent variables.

\section{Discussion}\label{sec:discussion}

\noindent This work has proposed a novel perspective on variational likelihood minimization, in which the distribution-valued first variation is directly fed into a formal version of a classical Wasserstein Fisher-Rao gradient flow. This approach yields simple closed-form expressions for tangent-space projections on statistical manifolds,  leading to particularly tractable numerical schemes. The Gaussian-mixture construction provides a concrete illustration of this principle.\\

\noindent As a particular case of our statistical manifold representation, a construction based on Gaussian-mixtures was proposed to implement numerical experiments, illustrating the applicability of the method to infer intensities of Poisson and Hawkes processes models. Furthermore, our algorithm merges nearby Gaussian components, yielding parsimonious representations of the estimated intensities.  On the statistical side, we obtain convergence of the empirical generalized gradient under suitable assumptions. These results lead to a conditional consistency theorem for projected equilibria.\\

\noindent  A natural starting point for further work is the joint estimation of a non-parametric triggering kernel and the background intensity, with particular attention to the identifiability of these components. In addition, the extension to time-dependent background intensities, under suitable assumptions on their temporal variation, also deserves investigation. In a different setting, the determination of convergence rates for the proposed estimators remains an interesting question to be addressed in future work.

\appendix

\section{Stationarity and prehistory estimates}\label{app:uniform-law-large-numbers}

\noindent This appendix supplies the stationary process used to define $\mathcal A$ in Subsection~\ref{subsec:consistency-deterministic-limit} and the coupling needed to recover that limit from the empty-history observations. We first verify stationarity, ergodicity and finite mean event rate. We then separate the events generated by pre-zero immigrants and bound their remaining contribution. These bounds are used in Subsection~\ref{appsubsec:empty-history-comparison} to prove that the two empirical operators have the same limit.

\noindent We work on $\mathcal Z=\mathbb R^d$ with the mark distribution and triggering kernel of Subsection~\ref{subsec:hawkes-etas}, and with branching ratio $\mathfrak n<1$ from \eqref{eq:etas-subcriticality}. Stationarity refers only to time translations: for $a\in\mathbb R$, set $\vartheta_a\boldsymbol\eta^\circ[A\times C\times D]=\boldsymbol\eta^\circ[(A+a)\times C\times D]$. No spatial stationarity is imposed. The following proposition provides the ergodic shift and finite unit-block counts required in Subsection~\ref{subsec:consistency-deterministic-limit}.

\begin{proposition}\label{prop:stationary-etas-regime}
Under the subcritical ETAS conditions of Subsection~\ref{subsec:hawkes-etas} and $\int_{\mathcal Z}\rho_\star\,dx<\infty$, the model admits a stationary version that is mixing under time translations. In particular, $S:=\vartheta_1$ is ergodic and
\begin{align}\label{eq:stationary-count-bound}
\mathbb E_\star^\circ[\eta^\circ[(0,1]\times\mathcal Z]]\leq\frac{\int_{\mathcal Z}\rho_\star(x)\,dx}{1-\mathfrak n}<\infty.
\end{align}
\end{proposition}

\begin{proof}
The Hawkes--Oakes construction uses Poisson immigrants with intensity $dt\,\rho_\star(x)\,dx$. Each immigrant receives an independent mark with density $q_{\mathrm{GR}}$ and generates an independent branching cluster \cite{hawkes1974cluster}. A parent with mark $r$ has a Poisson family of direct descendants with space-time intensity $\nu(s,z,r)\,ds\,dz$, so its expected number of children is $\kappa(r)$. The independent marks of new children make the expected size of generation $j$ equal to $\mathfrak n^j$. Consequently the expected total size of one cluster is
\[
\sum_{j=0}^{\infty}\mathfrak n^j=\frac1{1-\mathfrak n}.
\]
It is finite, so clusters are finite almost surely. Placing immigrants on the whole time axis and integrating their expected cluster contributions over immigrant times gives \eqref{eq:stationary-count-bound}; on the full spatial domain the displayed inequality is an equality. In particular, there are finitely many events in every bounded time interval almost surely. Time translation preserves the immigrant law and the relative cluster law, giving stationarity, and summing the offspring intensities gives the stipulated Hawkes compensator. For mixing, regard each immigrant as marked by its entire independent relative cluster. This marked Poisson immigrant process is mixing under time translations, and the measurable cluster-superposition map commutes with those translations. Its image is therefore mixing, so the unit-time shift of the Hawkes process is ergodic; see \cite{westcott1971mixing,bremaud1996stability,zhuang2013stability}.
\end{proof}

\noindent Finite mean rate also gives $\int_{\mathcal Z}\mathbb E_\star^\circ[B_0^\circ(x)]\,dx<\infty$, by stationarity and the compensator identity. Hence $B_0^\circ(x)$ is finite almost surely for almost every $x$, as required in \eqref{eq:population-score-consistency}. The predictable set where $B_t^\circ(x)=\infty$ has zero compensator measure and therefore contains no observed events almost surely. Thus the stationary event formulas use finite excitation values as well as positive candidate denominators.

\subsection{Contribution of the initial history}\label{appsubsec:prehistory-estimates}

\noindent We now couple the stationary and empty-history processes so that their difference is expressed entirely through families whose immigrants arrived before zero. Retain only the clusters whose immigrants occur after zero to obtain the observed process. On positive times,
\begin{align}\label{eq:prehistory-coupling}
\eta^\circ=\eta+\eta^{\mathrm{pre}},\qquad B_t^\circ=B_t+R_t,\qquad R_t\geq0,
\end{align}
where $\eta^{\mathrm{pre}}$ consists of post-zero events whose immigrant ancestor occurred before zero. The field $R_t$ is the excitation generated by all events in those pre-zero immigrant families that occurred before $t$, including events before zero. Families are classified by their immigrant ancestor, not by the birth times of individual descendants. The two collections of families are independent. Write $\mathbb E_\star$ for expectation on this coupling space, whose empty-history marginal is the original data law. In the joint past filtration, which reveals events and their family labels only as they occur, both excitation fields are predictable, $\eta$ has intensity $\lambda_\star(t,x)=\rho_\star(x)+B_t(x)$, and $\eta^{\mathrm{pre}}$ has intensity $R_t(x)$.

\noindent To bound the missing-history contribution, define
\[
\psi(t)=\sum_{j=1}^\infty\mathfrak n^j g^{*j}(t),\qquad \Psi(t)=\int_t^\infty\psi(s)\,ds.
\]
The $j$th generation has expected temporal density $\mathfrak n^j g^{*j}$ after integrating over spatial locations and marks. An immigrant at time $-u$ contributes at time $t>0$ through descendants of age $t+u$. Summing over generations and integrating over $u>0$ gives
\begin{align}\label{eq:prehistory-excitation-bound}
\int_{\mathcal Z}\mathbb E_\star[R_t(x)]\,dx
&\leq\bigg(\int_{\mathcal Z}\rho_\star(x)\,dx\bigg)\sum_{j=1}^\infty\mathfrak n^j\int_t^\infty g^{*j}(s)ds
=\bigg(\int_{\mathcal Z}\rho_\star(x)\,dx\bigg)\Psi(t).
\end{align}
The compensator of $\eta^{\mathrm{pre}}$ then yields
\begin{align}\label{eq:prehistory-count-bound}
\mathbb E_\star[\eta^{\mathrm{pre}}[(0,T]\times\mathcal Z]]
&=\int_0^T\int_{\mathcal Z}\mathbb E_\star[R_t(x)]\,dx\,dt \leq\bigg(\int_{\mathcal Z}\rho_\star(x)\,dx\bigg)\int_0^T\Psi(t)\,dt.
\end{align}
Since $\int_0^\infty\psi(s)\,ds=\mathfrak n/(1-\mathfrak n)$, the function $\Psi$ is bounded and tends to zero. In particular, $T^{-1}\int_0^T\Psi(t)\,dt$ tends to zero. These are the two estimates needed in the operator comparison: they control both the change in excitation at shared events and the additional events belonging to pre-zero families. No temporal first moment is required.

\section{Identification of the background intensity}\label{app:localization}

\noindent This appendix proves Proposition~\ref{prop:population-identification} from Subsection~\ref{subsec:consistency-deterministic-limit}. We first justify the stationary expectation defining $\mathcal A$ and its continuity in the candidate density. We then show that $\mathcal A(\rho)=0$ forces $\rho=\rho_\star$ when $\kappa_{FR}>0$. Continuity and uniqueness are used in Subsection~\ref{subsec:consistency-tangent-approximation} to turn a small limiting residual into convergence of the selected densities. Throughout, the model and candidate class are those of Section~\ref{sec:consistency}.\\

\noindent We begin with a bound for the full event integrand. It makes the stationary expectation finite and supplies the domination needed to pass to a limit in $\rho$. Factorization \eqref{eq:force-factorization} and Sobolev embedding give, for $\rho\in\mathcal K$, $\phi\in H^\beta$ and $b\geq0$,
\begin{align}\label{eq:full-integrand-bound}
\bigg|\frac{-\kappa_W\nabla\cdot(\rho\nabla\phi)(x)+\kappa_{FR}\rho(x)\phi(x)}{\rho(x)+b}\bigg|\leq C_0(1+|\nabla\log\rho(x)|)\|\phi\|_{H^\beta},
\end{align}
where $C_0=C_\beta(\kappa_W+\kappa_{FR})$. The random variable $Z$ in \eqref{eq:localized-integrability} is measurable: on a finite unit-block configuration, the event sum defining it is continuous in $\rho\in\mathcal K$, so its supremum equals the supremum over one countable dense subset of $\mathcal K$. Values on the null set of nonfinite configurations may be assigned arbitrarily. Consequently,
\begin{align}\label{eq:stationary-unit-operator-bound}
\sup_{\rho\in\mathcal K}\|\mathcal A_1^\circ(\rho)\|_{H^{-\beta}}\leq C_0Z+\kappa_{FR}C_\beta M.
\end{align}
The right-hand side is integrable by \eqref{eq:localized-integrability}. Taking expectations through the dual pairings therefore defines $\mathcal A(\rho)\in H^{-\beta}$.\\

\noindent We next justify the score representation used for identification. The score in \eqref{eq:population-score-consistency} is locally bounded because $|\mathfrak s_\rho|\leq|\rho_\star-\rho|/\rho$, and it satisfies
\begin{align}\label{eq:weighted-population-score}
\rho|\mathfrak s_\rho|\leq|\rho_\star-\rho|,\qquad |\nabla\rho|\,|\mathfrak s_\rho|\leq\rho_\star|\nabla\log\rho|+|\nabla\rho|.
\end{align}
By the compensator identity, first for nonnegative truncations and then by monotone convergence,
\begin{align}\label{eq:true-background-log-derivative-integrability}
\int_{\mathcal Z}\rho_\star|\nabla\log\rho|\,dx\leq\mathbb E_\star^\circ\bigg[\int_{(0,1]\times\mathcal Z}|\nabla\log\rho(x)|\,\eta^\circ(dt,dx)\bigg]\leq\mathbb E_\star^\circ[Z].
\end{align}
Together with $\rho\in W^{1,1}$, these estimates give \[\int\rho|\mathfrak s_\rho|\,dx<\infty,\qquad\int|\nabla\rho|\,|\mathfrak s_\rho|\,dx<\infty.\] Since $\nabla\cdot(\rho\nabla\phi)=\nabla\rho\cdot\nabla\phi+\rho\Delta\phi$, the integral in \eqref{eq:limiting-generalized-gradient} is absolutely convergent for every $\phi\in H^\beta$.

\noindent Also, $\nabla\cdot(\rho\nabla\phi)\in L^1$ and its integral is zero. Indeed, testing with a smooth cutoff equal to one on $K_R$ bounds the boundary term by $C R^{-1}\|\rho\|_{L^1}\|\nabla\phi\|_\infty$, which tends to zero. Applying the stationary compensator to \eqref{eq:consistency-stationary-mean} gives
\begin{align}\label{eq:population-compensator-representation}
\langle\mathcal A(\rho),\phi\rangle
={}&\int_{\mathcal Z}\mathbb E_\star^\circ\bigg[\frac{\lambda_\star^\circ(0,x)}{\rho(x)+B_0^\circ(x)}\bigg]\big[-\kappa_W\nabla\cdot(\rho\nabla\phi)(x)+\kappa_{FR}\rho(x)\phi(x)\big]\,dx
-\kappa_{FR}\int_{\mathcal Z}\rho\phi\,dx.
\end{align}
The interchanges are justified by \eqref{eq:full-integrand-bound} and \eqref{eq:localized-integrability}. Because the divergence integrates to zero, subtracting the last term replaces the expected ratio by $\mathfrak s_\rho$, giving \eqref{eq:limiting-generalized-gradient}.\\

\noindent For continuity, let $\rho_n,\rho\in\mathcal K$ with $d_{\mathcal K}(\rho_n,\rho)\to0$. The stationary unit block contains finitely many events with finite excitation values almost surely. Since $\rho_n\to\rho$ in $H^\beta$ and $\rho$ is strictly positive, the candidate intensities at these locations have a common positive lower bound for all sufficiently large $n$. Proposition~\ref{prop:empirical-generalized-gradient} therefore gives
\[
\|\mathcal A_1^\circ(\rho_n)-\mathcal A_1^\circ(\rho)\|_{H^{-\beta}}\longrightarrow0\qquad\text{almost surely}.
\]
The differences are dominated by twice the integrable bound in \eqref{eq:stationary-unit-operator-bound}; their norms are measurable by separability of $H^\beta$. Thus dominated convergence and the definition of $\mathcal A$ yield
\[
\|\mathcal A(\rho_n)-\mathcal A(\rho)\|_{H^{-\beta}}\leq\mathbb E_\star^\circ\big[\|\mathcal A_1^\circ(\rho_n)-\mathcal A_1^\circ(\rho)\|_{H^{-\beta}}\big]\longrightarrow0.
\]
This proves the continuity asserted in Proposition~\ref{prop:population-identification}.\\

\noindent It remains to show why the equation $\mathcal{A}(\rho)=0$  identifies the true background. The score factorization makes $\mathcal A(\rho_\star)=0$ immediate, but the converse requires ruling out nonzero scores annihilated by the weighted differential operator. We first obtain the local regularity needed to test the resulting equation and then use a cutoff argument on the whole spatial domain.\\

\noindent Suppose $\kappa_{FR}>0$ and $\mathcal A(\rho)=0$, and write $v=\mathfrak s_\rho$. Equations \eqref{eq:weighted-population-score}--\eqref{eq:true-background-log-derivative-integrability} give
\begin{align}\label{eq:identification-weighted-integrability}
\int_{\mathcal Z}\rho|v|\,dx<\infty,\qquad\int_{\mathcal Z}|\nabla\rho|\,|v|\,dx<\infty.
\end{align}
If $\kappa_W=0$, the equation is $\kappa_{FR}\rho v=0$, so $v=0$ almost everywhere. For $\kappa_W>0$, testing \eqref{eq:limiting-generalized-gradient} against $C_c^\infty$ gives
\begin{align}\label{eq:identification-local-equation}
\nabla\cdot(\rho\nabla v)=\frac{\kappa_{FR}}{\kappa_W}\rho v
\end{align}
in distributions, where the left-hand side acts on a test $\phi$ as $\int v\nabla\cdot(\rho\nabla\phi)\,dx$.\\

\noindent To justify the tests used below, set $u=\rho v$ and $b=\nabla\log\rho$. The score is locally bounded, and $\rho\in H^\beta$ with $\beta>d/2+2$ has a positive $C^2$ representative. Hence $u$ and $b$ are locally bounded. Equation \eqref{eq:identification-local-equation} becomes $\Delta u=\nabla\cdot(bu)+(\kappa_{FR}/\kappa_W)u$. For $\zeta\in C_c^\infty$, the function $w=\zeta u\in L^2$ satisfies $\Delta w=\nabla\cdot F+f$, where
\[
F=\zeta bu+2u\nabla\zeta,\qquad f=(\kappa_{FR}/\kappa_W)\zeta u-u b\cdot\nabla\zeta-u\Delta\zeta.
\]
Both $F$ and $f$ are compactly supported and in $L^2$. On frequencies with $|\xi|\geq1$, the Fourier identity bounds $|\xi|$ times the absolute transform of $w$ by the absolute transform of $F$ plus $|\xi|^{-1}$ times that of $f$; frequencies with $|\xi|<1$ are controlled by $\|w\|_{L^2}$. Plancherel's identity gives
\[
\|w\|_{H^1}\leq C\big(\|w\|_{L^2}+\|F\|_{L^2}+\|f\|_{L^2}\big).
\]
Since the cutoff is arbitrary, $u\in H^1_{\mathrm{loc}}$. On any compact set, $1/\rho$ and its first derivatives are bounded, so $v=u/\rho\in H^1_{\mathrm{loc}}$. This regularity is derived only at a zero of $\mathcal A$ and does not require differentiating the excitation field.\\

\noindent Let $\chi\in C_c^\infty$ be nonnegative and $F_\varepsilon(a)=\sqrt{a^2+\varepsilon^2}$. For fixed $\varepsilon>0$, $F_\varepsilon'$ is bounded and Lipschitz, so $\chi F_\varepsilon'(v)$ is a compactly supported $H^1$ function. On a bounded open set containing its support, the weak equation extends continuously from smooth tests to $H^1_0$ tests. Approximation therefore justifies using $\chi F_\varepsilon'(v)$. The chain rule and $F_\varepsilon''\geq0$ give
\begin{align*}
\kappa_W\int F_\varepsilon(v)\nabla\cdot(\rho\nabla\chi)\,dx&=\kappa_W\int\chi\rho F_\varepsilon''(v)|\nabla v|^2\,dx+\kappa_{FR}\int\chi\rho vF_\varepsilon'(v)\,dx\\
&\geq\kappa_{FR}\int\chi\rho vF_\varepsilon'(v)\,dx.
\end{align*}
On the support of $\chi$, $F_\varepsilon(v)\leq|v|+1$ for $\varepsilon\leq1$, and $0\leq vF_\varepsilon'(v)\leq|v|$. Dominated convergence as $\varepsilon$ decreases to zero yields
\begin{align}\label{eq:identification-absolute-value}
\kappa_{FR}\int\chi\rho|v|\,dx\leq\kappa_W\int|v|\nabla\cdot(\rho\nabla\chi)\,dx.
\end{align}
Choose $\chi_R(x)=\chi_1(x/R)$, where $\chi_1\in C_c^\infty$, $0\leq\chi_1\leq1$, and $\chi_1$ is one on $K_1$ and zero outside $K_2$. With $C_1=\|\nabla\chi_1\|_\infty$ and $C_2=\|\Delta\chi_1\|_\infty$, the right-hand side is bounded by
\[
\kappa_W\bigg(\frac{C_2}{R^2}\int\rho|v|\,dx+\frac{C_1}{R}\int|\nabla\rho|\,|v|\,dx\bigg).
\]
This tends to zero by \eqref{eq:identification-weighted-integrability}, while dominated convergence applies on the left. Hence $\int\rho|v|\,dx=0$, and strict positivity of $\rho$ gives $v=0$ almost everywhere.\\

\noindent In either case, the reciprocal expectation in \eqref{eq:population-score-consistency} is strictly positive for almost every $x$, since $\rho(x)>0$ and $B_0^\circ(x)<\infty$ almost surely there. Thus $v=0$ implies $\rho=\rho_\star$ almost everywhere, and their continuous representatives agree everywhere. This completes the proof of Proposition~\ref{prop:population-identification}.

\section{Uniform convergence of the empirical generalized gradient}\label{app:uniform-convergence}

\noindent This appendix proves Proposition~\ref{prop:uniform-generalized-gradient-convergence} from Subsection~\ref{subsec:consistency-uniform-convergence}. We need an approximation that remains valid at a data-dependent candidate and a bound that controls the tangent-approximation error in Theorem~\ref{thm:projected-consistency}. We first verify compactness of the test-function class and obtain uniform convergence for the stationary event blocks. We then use the coupling of Appendix~\ref{app:uniform-law-large-numbers} to transfer the limit to the empty-history observations, and finish with the uniform $H^{-\beta}$ bound. The model, candidate class and integrability assumptions are those of Section~\ref{sec:consistency}.

\subsection{Compactness of the test class}\label{appsubsec:test-functions}

\noindent The uniform ergodic argument requires a compact parameter space containing both the candidate densities and the test functions. Compactness of $\mathcal K$ is assumed; here we verify the compact inclusion for $\mathcal X^{\beta+r}$ asserted in Subsection~\ref{subsec:consistency-uniform-convergence}. Fix $r>0$. A bounded sequence in $\mathcal X^{\beta+r}$ has uniformly bounded $H^{\beta+r}$ norm and satisfies
\[
\int_{\{|x|>R\}}|\phi_n(x)|^2\,dx\leq R^{-2}\||x|\phi_n\|_{L^2}^2.
\]
Rellich compactness on each ball and a diagonal argument give a subsequence converging locally in $L^2$. The displayed tail bound makes it Cauchy in $L^2(\mathbb R^d)$. Interpolation gives, for two terms of that subsequence,
\[
\|\phi_j-\phi_k\|_{H^\beta}\leq\|\phi_j-\phi_k\|_{L^2}^{r/(\beta+r)}\|\phi_j-\phi_k\|_{H^{\beta+r}}^{\beta/(\beta+r)}.
\]
The second factor is uniformly bounded, so the subsequence is Cauchy in $H^\beta$. This proves compactness of the inclusion. Separately, $C_c^\infty\subset\mathcal X^{\beta+r}$ and density of $C_c^\infty$ in $H^\beta$ show that $\mathcal X^{\beta+r}$ is dense in $H^\beta$.\\

\noindent We now prove uniform convergence for $\mathcal A_T^\circ$, using the stationary process of Proposition~\ref{prop:stationary-etas-regime}. Let $\mathbb B_r$ be the unit ball of $\mathcal X^{\beta+r}$ and put $\mathsf P=\mathcal K\times\overline{\mathbb B_r}^{\,H^\beta}$, with metric $d((\rho,\phi),(\gamma,\psi))=d_{\mathcal K}(\rho,\gamma)+\|\phi-\psi\|_{H^\beta}$. The preceding subsection makes $\mathsf P$ compact. Define the full unit-block contributions by
\begin{align}\label{eq:ulln-integrand}
H_{\rho,\phi}^\circ(t,x)&=\frac{-\kappa_W\nabla\cdot(\rho\nabla\phi)(x)+\kappa_{FR}\rho(x)\phi(x)}{\rho(x)+B_t^\circ(x)},\nonumber\\
\Xi_j^\circ(\rho,\phi)&=\int_{(j,j+1]\times\mathcal Z}H_{\rho,\phi}^\circ(t,x)\,\eta^\circ(dt,dx).
\end{align}
Stationarity gives $\Xi_j^\circ=\Xi_0^\circ\circ S^j$, where $S=\vartheta_1$ is measure preserving and ergodic.\\

\noindent We apply the uniform ergodic theorem for random continuous functions, Theorem~2.1 of Hansen \cite{hansen2012proofs}; see also \cite{rangarao1962relations,hansen1982large}. The stationary unit block contains finitely many events with finite excitation values almost surely. On each such realization, $\Xi_0^\circ$ is a finite sum of continuous functions of $(\rho,\phi)\in\mathsf P$: the candidate denominators are positive, and the required point evaluations are continuous by Sobolev embedding. Pointwise measurability and separability of $C(\mathsf P)$ make $\Xi_0^\circ$ a measurable $C(\mathsf P)$-valued random element, after assigning values on the null exceptional set. Furthermore, \eqref{eq:full-integrand-bound} gives
\[
\sup_{z\in\mathsf P}|\Xi_0^\circ(z)|\leq C_0Z,\qquad\mathbb E_\star^\circ[Z]<\infty.
\]
For each $z\in\mathsf P$, continuity makes $\sup_{d(z,z')<\delta}|\Xi_0^\circ(z)-\Xi_0^\circ(z')|$ tend to zero almost surely as $\delta\downarrow0$. This measurable supremum is bounded by $2C_0Z$, so dominated convergence verifies the first-moment continuity required by the cited theorem. It follows that
\begin{align}\label{eq:stationary-block-ulln}
\lim_{n\to\infty}\sup_{z\in\mathsf P}\bigg|\frac1n\sum_{j=0}^{n-1}\Xi_j^\circ(z)-\mathbb E_\star^\circ[\Xi_0^\circ(z)]\bigg|=0\qquad\text{almost surely}.
\end{align}
By the definition of $\mathcal A$,
\[
\langle\mathcal A_n^\circ(\rho)-\mathcal A(\rho),\phi\rangle=\frac1n\sum_{j=0}^{n-1}\Xi_j^\circ(\rho,\phi)-\mathbb E_\star^\circ[\Xi_0^\circ(\rho,\phi)].
\]
The pairings are $H^\beta$-continuous, so taking the supremum over $\mathsf P$ is equivalent to taking the supremum over $\rho\in\mathcal K$ and the original unit ball $\mathbb B_r$. This proves the stationary uniform limit at integer horizons.\\

\noindent For arbitrary $T$, put $n=\lfloor T\rfloor\geq1$ and $Z_j=Z\circ S^j$. The deterministic mass term cancels when comparing $\mathcal A_T^\circ$ with $\mathcal A_n^\circ$. The difference of the event averages, uniformly over $\mathsf P$, is at most
\[
\frac{C_0}{T}Z_n+C_0\bigg|\frac1T-\frac1n\bigg|\sum_{j=0}^{n-1}Z_j.
\]
The ergodic theorem gives $n^{-1}\sum_{j=0}^{n-1}Z_j\to\mathbb E_\star^\circ[Z]$ almost surely. On that probability-one event,
\[
\frac{Z_n}{n}=\frac{n+1}{n}\bigg(\frac1{n+1}\sum_{j=0}^{n}Z_j\bigg)-\frac1n\sum_{j=0}^{n-1}Z_j\longrightarrow0.
\]
Since $n/T\to1$ and $|1/T-1/n|\leq1/(nT)$, both terms in the preceding bound vanish. Thus
\begin{align}\label{eq:stationary-uniform-gradient-localized}
\lim_{T\to\infty}\sup_{\rho\in\mathcal K}\|\mathcal A_T^\circ(\rho)-\mathcal A(\rho)\|_{\mathcal X^{-(\beta+r)}}=0\qquad\text{almost surely}.
\end{align}
For a single fixed $\phi\in H^\beta$, the scalar ergodic theorem and the same partial-block bound also give the fixed-test convergence used in Subsection~\ref{subsec:consistency-deterministic-limit}.

\subsection{Comparison with the observed process}\label{appsubsec:empty-history-comparison}

\noindent The stationary limit must now be transferred to the operator computed from the actual observations. On the coupling of Subsection~\ref{appsubsec:prehistory-estimates}, we prove the stronger comparison
\begin{align}\label{eq:positive-stationary-gradient-comparison}
\lim_{T\to\infty}\mathbb E_\star\bigg[\sup_{\rho\in\mathcal K}\|\mathcal A_T(\rho)-\mathcal A_T^\circ(\rho)\|_{H^{-\beta}}\bigg]=0.
\end{align}
The argument uses a fixed spatial ball to control the denominators and the integrability assumption to control the events outside that ball. This localization is needed for the comparison, not for the stationary uniform ergodic argument above.\\

\noindent First, for $R\geq1$, compactness of $\mathcal K$ and Sobolev embedding make $(\rho,x)\mapsto\rho(x)$ and $(\rho,x)\mapsto\nabla\rho(x)$ continuous on $\mathcal K\times K_R$. Strict positivity therefore gives
\begin{align}\label{eq:local-density-lower-bound}
c_R:=\inf_{\rho\in\mathcal K,\,x\in K_R}\rho(x)>0,\qquad L_R:=\sup_{\rho\in\mathcal K,\,x\in K_R}|\nabla\log\rho(x)|<\infty.
\end{align}
Neither constant is required to be uniform in $R$. For the tails, define
\[
Z_R:=\sup_{\rho\in\mathcal K}\int_{(0,1]\times K_R^c}(1+|\nabla\log\rho(x)|)\,\eta^\circ(dt,dx).
\]
As for $Z$, the supremum can be taken over a countable dense subset of $\mathcal K$, so $Z_R$ is measurable. Each finite unit-block configuration is contained in some ball, making $Z_R$ eventually zero on that realization. Since $0\leq Z_R\leq Z$, dominated convergence yields
\begin{align}\label{eq:localized-tail-zero}
\lim_{R\to\infty}\mathbb E_\star^\circ[Z_R]=0.
\end{align}

\noindent Keep $R$ fixed and set
\[
H_{\rho,\phi}(t,x)=\frac{-\kappa_W\nabla\cdot(\rho\nabla\phi)(x)+\kappa_{FR}\rho(x)\phi(x)}{\rho(x)+B_t(x)}.
\]
The deterministic mass terms cancel, and \eqref{eq:prehistory-coupling} gives
\begin{align*}
\langle\mathcal A_T(\rho)-\mathcal A_T^\circ(\rho),\phi\rangle
&=\frac1T\int_{(0,T]\times K_R}(H_{\rho,\phi}-H_{\rho,\phi}^\circ)(t,x)\,\eta(dt,dx)\\
&-\frac1T\int_{(0,T]\times K_R}H_{\rho,\phi}^\circ(t,x)\,\eta^{\mathrm{pre}}(dt,dx)\\
&+\frac1T\int_{(0,T]\times K_R^c}H_{\rho,\phi}(t,x)\,\eta(dt,dx)\\
&-\frac1T\int_{(0,T]\times K_R^c}H_{\rho,\phi}^\circ(t,x)\,\eta^\circ(dt,dx).
\end{align*}
The first term accounts for the changed excitation at shared events, the second for events absent from the observed process, and the last two for the spatial tails.

\noindent For $x\in K_R$ and $\|\phi\|_{H^\beta}\leq1$, factorization \eqref{eq:force-factorization} gives
\begin{align*}
|H_{\rho,\phi}-H_{\rho,\phi}^\circ|&\leq C_0(1+L_R)\frac{\rho(x)R_t(x)}{(\rho(x)+B_t(x))(\rho(x)+B_t(x)+R_t(x))}\\
&\leq C_0(1+L_R)\frac{R_t(x)}{c_R+B_t(x)+R_t(x)}.
\end{align*}
Also $|H_{\rho,\phi}^\circ|\leq C_0(1+L_R)$ on $K_R$. With $D_R=\max\{1,\sup_{x\in K_R}\rho_\star(x)/c_R\}$,
\[
\frac{\lambda_\star(t,x)}{c_R+B_t(x)+R_t(x)}\leq D_R,\qquad x\in K_R.
\]
In the joint past filtration, $R_t$ is predictable, and the compensator of $\eta$ is $\lambda_\star(t,x)\,dt\,dx$. Taking absolute values, then the supremum over $\rho$ and unit $H^\beta$ tests, bounds the expected contribution of the first two terms by
\[
\frac{C_0(1+L_R)}{T}\bigg(D_R\int_0^T\int_{K_R}\mathbb E_\star[R_t(x)]\,dx\,dt+\mathbb E_\star[\eta^{\mathrm{pre}}[(0,T]\times K_R]]\bigg).
\]
Equations \eqref{eq:prehistory-excitation-bound}--\eqref{eq:prehistory-count-bound} bound this expression by $C_R T^{-1}\int_0^T\Psi(t)\,dt$, where one may take
\[
C_R=C_0(1+L_R)(D_R+1)\int_{\mathcal Z}\rho_\star(x)\,dx.
\]

\noindent For the two tail terms, \eqref{eq:full-integrand-bound} and $\eta\leq\eta^\circ$ give the uniform bound
\[
\frac{2C_0}{T}\sup_{\rho\in\mathcal K}\int_{(0,T]\times K_R^c}(1+|\nabla\log\rho(x)|)\,\eta^\circ(dt,dx)\leq\frac{2C_0}{T}\sum_{j=0}^{\lceil T\rceil-1}Z_R\circ S^j.
\]
For $T\geq1$, its expectation is at most $4C_0\mathbb E_\star^\circ[Z_R]$. Combining the local and tail estimates yields
\begin{align}\label{eq:positive-stationary-gradient-bound}
\mathbb E_\star\bigg[\sup_{\rho\in\mathcal K}\|\mathcal A_T(\rho)-\mathcal A_T^\circ(\rho)\|_{H^{-\beta}}\bigg]\leq\frac{C_R}{T}\int_0^T\Psi(t)\,dt+4C_0\mathbb E_\star^\circ[Z_R].
\end{align}
For fixed $R$, the first term tends to zero because $\Psi$ is bounded and tends to zero. Taking the limit superior in $T$ and only then letting $R$ tend to infinity proves \eqref{eq:positive-stationary-gradient-comparison}. The comparison also holds in $\mathcal X^{-(\beta+r)}$ for every $r\geq0$, since the $\mathcal X^{\beta+r}$ norm dominates the $H^\beta$ norm. No temporal first moment or relation between $R$ and $T$ is needed.\\

\noindent We now combine the two limits and prove the stochastic bound in Proposition~\ref{prop:uniform-generalized-gradient-convergence}. The triangle inequality gives
\begin{align*}
\sup_{\rho\in\mathcal K}\|\mathcal A_T(\rho)-\mathcal A(\rho)\|_{\mathcal X^{-(\beta+r)}}\leq{}&\sup_{\rho\in\mathcal K}\|\mathcal A_T^\circ(\rho)-\mathcal A(\rho)\|_{\mathcal X^{-(\beta+r)}}
+\sup_{\rho\in\mathcal K}\|\mathcal A_T(\rho)-\mathcal A_T^\circ(\rho)\|_{\mathcal X^{-(\beta+r)}}.
\end{align*}
The first term tends to zero almost surely by \eqref{eq:stationary-uniform-gradient-localized}; the second tends to zero in mean, hence in probability, by \eqref{eq:positive-stationary-gradient-comparison}. This proves \eqref{eq:uniform-generalized-gradient-convergence} on the coupling space. Since $\mathcal A_T$ depends only on the observed process and $\mathcal A$ is deterministic, the quantity on the left-hand side of \eqref{eq:uniform-generalized-gradient-convergence} has the same distribution under the coupling and the original data law. Consequently, the convergence in probability in \eqref{eq:uniform-generalized-gradient-convergence} also holds under the original data law.\\

\noindent For the remaining bound, \eqref{eq:full-integrand-bound} and $\eta\leq\eta^\circ$ give
\[
\sup_{\rho\in\mathcal K}\|\mathcal A_T(\rho)\|_{H^{-\beta}}\leq\frac{C_0}{T}\sum_{j=0}^{\lceil T\rceil-1}Z\circ S^j+\kappa_{FR}C_\beta M.
\]
For $T\geq1$, the expectation is bounded by $2C_0\mathbb E_\star^\circ[Z]+\kappa_{FR}C_\beta M$, uniformly in $T$. Markov's inequality proves \eqref{eq:uniform-empirical-gradient-bound}. All suprema used here are measurable: the candidate class and test spaces are separable, and the maps $(\rho,\phi)\mapsto\langle\mathcal A_T(\rho),\phi\rangle$ and $(\rho,\phi)\mapsto\langle\mathcal A(\rho),\phi\rangle$ are continuous by Proposition~\ref{prop:empirical-generalized-gradient} and Appendix~\ref{app:localization}. This completes the proof of Proposition~\ref{prop:uniform-generalized-gradient-convergence}. Its two conclusions are exactly the two bounds used in \eqref{eq:limiting-gradient-at-empirical-equilibrium}.

\section{Gaussian classes and spatial integrability}\label{app:gaussian-class}

\noindent This appendix proves the Gaussian compatibility result, Proposition~\ref{prop:gaussian-localization}, from Subsection~\ref{subsec:consistency-gaussian-classes}. We first show that the allowed mixing measures generate a compact class of positive regular densities and verify the logarithmic-derivative bound. We then give sufficient conditions on the true background and triggering kernel for the stationary spatial moment required in \eqref{eq:stationary-spatial-first-moment}. These calculations verify the analytical assumptions for the Gaussian class; they do not establish tangent approximation or the existence of empirical equilibria.\\

\noindent We use the bandwidth, center bound and mass interval fixed in Proposition~\ref{prop:gaussian-localization}. The allowed mixing measures have supports in a fixed compact set and masses in $[a,b]$, so they form a weakly compact set. To transfer compactness to the density metric, let $\xi_n$ converge weakly to an allowed $\xi$. The map $m\mapsto\chi_\sigma(\cdot;m)$ is continuous into the Banach space $H^\beta\cap W^{1,1}$ with its sum norm, hence uniformly continuous on the compact center set. For each $\varepsilon>0$, a finite partition of unity gives an approximation $\sum_{j=1}^J a_j(m)f_j$, with continuous scalar functions $a_j$ and fixed elements $f_j$ of that Banach space, whose error is at most $\varepsilon$, uniformly in $m$.\\

\noindent Integrating this error against either $\xi_n$ or $\xi$ gives a norm error at most $b\varepsilon$. For the finite approximation, weak convergence makes each coefficient $\int a_j(m)\,\xi_n(dm)$ converge to $\int a_j(m)\,\xi(dm)$. Thus the limit superior of the norm distance between the Gaussian convolutions is at most $2b\varepsilon$. Letting $\varepsilon$ decrease to zero proves convergence in $H^\beta\cap W^{1,1}$, hence in $d_{\mathcal K}$. The image of the mixing-measure class is therefore compact. Translation invariance and the triangle inequality give
\[
\|\rho\|_{H^\beta}+\|\rho\|_{W^{1,1}}\leq b\big(\|\chi_\sigma(\cdot;0)\|_{H^\beta}+\|\chi_\sigma(\cdot;0)\|_{W^{1,1}}\big).
\]
Since the Gaussian is strictly positive and $\xi[\mathbb R^d]\geq a>0$, each convolution is strictly positive everywhere. This verifies the class conditions in \eqref{eq:admissible-density-class}.\\

\noindent To check event integrability, differentiate under the mixing integral to obtain
\[
\nabla\log\rho(x)=-\frac{x}{\sigma^2}+\frac1{\sigma^2}\frac{\int m\chi_\sigma(x;m)\,\xi(dm)}{\int\chi_\sigma(x;m)\,\xi(dm)}.
\]
The ratio is a weighted mean of centers of norm at most $D$, proving \eqref{eq:gaussian-log-derivative-bound}. Consequently $Z$ is bounded by the stationary event integral of $1+(|x|+D)/\sigma^2$, and \eqref{eq:stationary-spatial-first-moment} implies \eqref{eq:localized-integrability}. This completes the proof of Proposition~\ref{prop:gaussian-localization}.\\

\noindent We finish by deriving the stationary spatial moment from assumptions on the background and the offspring displacements, rather than imposing it directly on the stationary process. Under the model assumptions of Proposition~\ref{prop:stationary-etas-regime}, suppose
\begin{align}\label{eq:sufficient-spatial-moments}
\int_{\mathcal Z}(1+|x|)\rho_\star(x)\,dx<\infty,\qquad \int_{M_u}^\infty\kappa(r)q_{\mathrm{GR}}(r)\int_{\mathbb R^d}|z|f(z;r)\,dz\,dr<\infty.
\end{align}
Write \[m_0=\int_{\mathcal Z}\rho_\star(x)\,dx,\qquad m_1=\int_{\mathcal Z}|x|\rho_\star(x)\,dx,\] and set \[h(z)=\int_{M_u}^\infty\kappa(r)q_{\mathrm{GR}}(r)f(z;r)\,dr,\qquad a_1=\int_{\mathbb R^d}|z|h(z)\,dz.\] Then $\int h=\mathfrak n<1$. For generation $j$, integrating over immigrant times removes the normalized temporal density, and independence of the marks attached to new events gives the mean spatial density $h^{*j}*\rho_\star$; for $j=0$ this means $\rho_\star$. The stationary mean spatial event density is the nonnegative sum of these generation contributions.

\noindent The $j$th term has mass $\mathfrak n^j m_0$. For $j\geq1$, write a descendant's position as $x+z_1+\cdots+z_j$. Integrating $|x+z_1+\cdots+z_j|\leq|x|+\sum_{i=1}^j|z_i|$ against $\rho_\star(x)\,dx\prod_{i=1}^j h(z_i)\,dz_i$ bounds its first absolute moment by $\mathfrak n^j m_1+j\mathfrak n^{j-1}a_1m_0$. Tonelli's theorem permits summation without assuming the desired stationary moment in advance. The two geometric series therefore give
\begin{align}\label{eq:stationary-spatial-moment-bound}
\mathbb E_\star^\circ\bigg[\int_{(0,1]\times\mathcal Z}(1+|x|)\,\eta^\circ(dt,dx)\bigg]\leq\frac{m_0+m_1}{1-\mathfrak n}+\frac{m_0a_1}{(1-\mathfrak n)^2}<\infty.
\end{align}
Thus \eqref{eq:sufficient-spatial-moments} implies the moment condition used in Subsection~\ref{subsec:consistency-gaussian-classes}. Only normalization of the temporal kernel is used, not a temporal first moment.

\section{Pseudocode}\label{sec:pseudocode} The pseudocode for the algorithm is presented below

\begin{algorithm}[H]
\caption{{WFR-Hawkes Algorithm (Spatial Background \& Kernel Estimation)}}\label{alg:WFR_Hawkes}
\begin{algorithmic}[1]
\Require Seismic catalog $\mathcal{D} = \{(t_i, \mathbf{x}_i, M_i)\}_{i=1}^N$; spatial bandwidth $\sigma > 0$; number of initial Gaussian components $k$; maximum outer iterations $I_{\max}$; number of updates using WFR in each iteration $N_{\text{WFR}}$; convergence threshold $\varepsilon$; merging distance $\delta_{\text{dist}}$; weight pruning threshold $\delta_{\text{weight}}$
\Ensure Estimated spatial background intensity $\mu(\mathbf{x})$ and optimal self-excitation parameters $\boldsymbol{\theta}_H^*$
\State Initialize centers $\{\mathbf{m}_j\}_{j=1}^k$ via uniform spatial sampling over domain of $\mathcal{D}$
\State Set equal initial log-weights $\alpha_j \gets \ln(1/k)$ for $j = 1, \dots, k$
\State Initialize ETAS kernel parameters $\boldsymbol{\theta}_H^{(0)} = (A, \alpha, c, p, d, \gamma, q)$ 
\State $\text{iter} \gets 0$, $\Delta \mathcal{L} \gets 2\varepsilon$, $\mathcal{L}_{\text{prev}} \gets -\infty$
\While{$\text{iter} < I_{\max}$ \textbf{and} $\Delta \mathcal{L} > \varepsilon$}
    \State $\text{iter} \gets \text{iter} + 1$
    \State \Comment{\textbf{Block 1: Background Density Optimization via WFR Flow}}
    \For{$s = 1$ \textbf{to} $N_{\text{wfr}}$}
        \State Compute spatial background density: $\mu(\mathbf{x}) \gets K_\mu \sum_{j=1}^k e^{\alpha_j} \mathcal{N}(\mathbf{x} \mid \mathbf{m}_j, \sigma^2 \mathbf{I})$
        \State Evaluate full conditional intensity: $\lambda(t, \mathbf{x}) \gets \mu(\mathbf{x}) + \sum_{t_j < t} g(t - t_j, \mathbf{x} - \mathbf{x}_j, M_j \mid \boldsymbol{\theta}_H)$
        \State With $\theta^{\text{iter}}$ fixed, compute log-likelihood gradient w.r.t. parameters $$\boldsymbol{\Theta}_{\text{bg}} = (\mathbf{m}_1, \dots, \mathbf{m}_k, \alpha_1, \dots, \alpha_k)$$
        \State Construct WFR Gram matrix $G(\boldsymbol{\Theta}_{\text{bg}})$ and gradient action vector $g_T(\boldsymbol{\Theta}_{\text{bg}})$
        \State Solve differential linear system: $G(\boldsymbol{\Theta}_{\text{bg}}) \, \dot{\boldsymbol{\Theta}}_{\text{bg}} = g_T(\boldsymbol{\Theta}_{\text{bg}})$
        \State Update centers $\mathbf{m}$ and log-weights $\boldsymbol{\alpha}$ using velocity $\dot{\boldsymbol{\Theta}}_{\text{bg}}$ via ADAM step
        \State \textbf{Component Merging and Pruning}
        \For{all pairs $(i, j)$ with $i \neq j$}
            \If{$\|\mathbf{m}_i - \mathbf{m}_j\|_2 < \delta_{\text{dist}}$}
                \State Merge component $j$ into $i$: $e^{\alpha_i} \gets e^{\alpha_i} + e^{\alpha_j}$ and recompute center $\mathbf{m}_i$
            \EndIf
        \EndFor
        \State Remove components $j$ where $e^{\alpha_j} < \delta_{\text{weight}}$ and re-normalize weights $\boldsymbol{\alpha}$
    \EndFor
    \State \Comment{\textbf{Block 2: Hawkes Self-Excitation Parameter Fitting}}
    \State Fix spatial background function $\mu(\mathbf{x})$
    \State Update kernel parameters: $\boldsymbol{\theta}_H^{(\text{iter})} \gets \arg\max_{\boldsymbol{\theta}_H} \ln \mathcal{L}(\mathcal{D} \mid \mu(\mathbf{x}), \boldsymbol{\theta}_H)$ 
    \State \Comment{\textbf{Convergence Check}}
    \State Compute total log-likelihood: $\mathcal{L}_{\text{total}} \gets \ln \mathcal{L}\left(\mathcal{D} \mid \mu^{(\text{iter})}, \boldsymbol{\theta}_H^{(\text{iter})}\right)$
    \State $\Delta \mathcal{L} \gets |\mathcal{L}_{\text{total}} - \mathcal{L}_{\text{prev}}|$
    \State $\mathcal{L}_{\text{prev}} \gets \mathcal{L}_{\text{total}}$
\EndWhile
\State \Return $\boldsymbol{\Theta}_{\text{bg}}$ and $\boldsymbol{\theta}_H^*$
\end{algorithmic}
\end{algorithm}

\noindent \textbf{Acknowledgements}\\
Arturo Jaramillo Gil was supported by
the grant CBF2023-2024-2088.

\bibliographystyle{plain}
\bibliography{references}

\end{document}